\documentclass[final,5p,times,twocolumn]{elsarticle}
\usepackage{framed,multirow}
\usepackage{amssymb}
\usepackage{latexsym}
\usepackage{nextpage}
\usepackage{natbib}
\usepackage{url}
\usepackage{xcolor}
\definecolor{newcolor}{rgb}{.8,.349,.1}
\usepackage{graphicx}
\usepackage{amsmath}
\usepackage{amssymb}
\usepackage{math}
\usepackage{enumitem}
\usepackage[]{algorithm2e}
\usepackage{placeins}
\usepackage{tabulary}
\usepackage{subfig}
\usepackage{caption}
\usepackage{tikz}
\usepackage[printonlyused,withpage]{acronym}
\usepackage[pagebackref=true,breaklinks=true,letterpaper=true,colorlinks,bookmarks=false]{hyperref}

\usepackage[final]{review}

\acrodef{MLE}{maximum likelihood estimation}
\acrodef{pdf}{probability distribution function}

\begin{document}

\begin{frontmatter}

\title{Unsupervised Performance Analysis of 3D Face Alignment with a Statistically Robust Confidence Test$^{*}$}
\tnotetext[t1]{This work is funded in part by the Multidisciplinary Institute in Artificial Intelligence (MIAI), Grenoble (ANR-19-P3IA-0003) and by the European Commission under the Horizon 2020 SPRING project (GA~871245).}
 \author[1]{Mostafa Sadeghi}
  \author[2]{Xavier Alameda-Pineda}
 \author[2]{Radu Horaud}
 \address[1]{Centre Inria Universit\'e de Lorraine, France}
 \address[2]{Centre Inria Universit\'e Grenoble Alpes, France}
             
\begin{abstract}
This paper addresses the problem of analysing the performance of 3D face alignment (3DFA), or facial landmark localization. This task is usually supervised, based on annotated datasets. Nevertheless, in the particular case of 3DFA, the annotation process is rarely error-free, which strongly biases the results. Alternatively, unsupervised performance analysis (UPA) is investigated. The core ingredient of the proposed methodology is the robust estimation of the rigid transformation between predicted landmarks and model landmarks. It is shown that the rigid mapping thus computed is affected neither by non-rigid facial deformations, due to variabilities in expression and in identity, nor by landmark localization errors, due to various perturbations. The guiding idea is to apply the estimated rotation, translation and scale to a set of predicted landmarks in order to map them onto a \textit{mathematical home} for the shape embedded in these landmarks (including possible errors). UPA proceeds as follows: (i) 3D landmarks are extracted from a 2D face using the 3DFA method under investigation; (ii) these landmarks are rigidly mapped onto a canonical (frontal) pose, and (iii) a statistically-robust confidence score is computed for each landmark. This allows to assess whether the mapped landmarks lie inside (inliers) or outside (outliers) a \textit{confidence volume}. An experimental evaluation protocol, that uses publicly available datasets and several 3DFA software packages associated with published articles, is described in detail. The results show that the proposed analysis is consistent with supervised metrics and that it can be used to measure the accuracy of both predicted landmarks and of automatically annotated 3DFA datasets, to detect errors and to eliminate them. Source code and supplemental materials for this paper are publicly available at \url{https://team.inria.fr/robotlearn/upa3dfa/}.
\end{abstract}

\begin{keyword}
Deep face alignment \sep 3D facial landmarks \sep Gaussian-uniform mixture \sep Student's t-distribution \sep robust statistical inference \sep rigid motion estimation \sep expectation-maximization algorithm \sep quaternion.
\end{keyword}

\end{frontmatter}

\section{Introduction}
\label{sec:introduction}

The problem of face alignment is the problem of facial landmark localization from a single RGB image. It is an important research topic as it provides input to a variety of tasks, e.g. face tracking, face recognition, expression recognition, visual speech processing, facial animation, etc., \citep{Escalera2018computational,Loy2019deep,wang2021deep}.
2D face alignment (2DFA) has been extensively studied for the last decades, yielding a plethora of methods and algorithms \citep{wu2018facial}. \addnote[srt]{2}{State of the art 2DFA is based on DNNs, e.g. \citep{deng2019joint},  or it combines DNN with differentiable optical-flow  and/or 3D-triangulation modules to supervise the location of 2D landmarks \citep{dong2018supervision,dong2020supervision}. }

In general, 2DFA yields poor performance in the presence of occlusions which occur in case of large poses induced by out-of-image-plane head rotations (self occlusions) as well as by the presence of various objects in the camera field of view, such as glasses, hair, hands or handheld objects. \addnote[2DFA-occlusions]{4}{We note however, that more recently there have been successful attempts to develop 2DFA that can deal with extreme poses and partial occlusions, e.g. \cite{wan2021arobust,wan2021brobust}. In particular, a recent method based on transformers, namely reference heatmap transformers (RHT) yields impressive 2DFA results \cite{wan2023precise}.
We also note that there is an increasing interest in 3DFA. 
3D facial landmarks embed both head-pose with six degrees of freedom and rich non-rigid deformation information . However, the caveat is that 3DFA training comes with additional difficulties, notably due to the fact that annotation should be carried out automatically: Indeed, accurate manual annotation of the $z$-coordinate (depth) is impractical if not impossible.}

%
%
%
\addnote[claims-0]{4}{This paper proposes Unsupervised Performance Analysis (UPA) for benchmarking 3DFA algorithms. At runtime, UPA-3DFA takes as input a set of 3D landmarks predicted from a 2D face and classifies these landmarks either as inliers or as outliers. The analysis can be applied to landmarks predicted by a trained DNN architecture, as well as to landmarks extracted with semi-automatic annotation techniques, e.g. \citep{gou2016shape,zhu2016face,deng2019menpo}. }

\addnote[claims-1]{4}{
The first contribution is the robust estimation of the rigid transformation (rotation, translation and scale) that maps the set of predicted 3D landmarks onto a frontally-viewed shape model. The guiding idea is to obtain a \textit{natural mathematical home} for the shape embedded in the predicted landmarks, i.e. \citep{kendall2009shape}. Indeed, the intrinsic shape properties, such as non-rigid deformations, are preserved under rotation, translation and scale. The challenge addressed in this paper is to estimate the rigid transformation that brings the landmarks associated with all the faces in the same frontal pose (or in the same coordinate frame) such that they can be directly compared.
}

\addnote[claims-2]{4}{
The second contribution is the construction of a \textit{statistical shape atlas}, on the following grounds.
Starting with 3D landmarks associated with a training dataset of face images with a large range of variabilities in pose, expression and identity,  the proposed robust rigid-mapping estimator is used to compute a \textit{statistical frontal landmark model} (SFL) and a \textit{confidence score}. SFL consists of a shape atlas, namely, ellipsoidal volumes computed from the posterior means and posterior covariances of the mapped landmarks. The size of each one of these ellipsoids is estimated in such a way that its \textit{inside} is an inlier volume with 99.7\% confidence. \textit{The built-in robust estimator downgrades the effect of large landmark errors that  are inherently present in the training dataset, thus preventing the inlier volumes to grow exaggeratedly large.}
The statistical shape atlas is thus conditioned by (i) the training dataset, (ii) the 3DFA method used to predict 3D landmarks, and (iii) the rigid-mapping parameters.
In practice, UPA proceeds as follows. Firstly, 3D landmarks are predicted with the 3DFA architecture under investigation. Secondly, the predicted landmarks are rigidly mapped onto the shape atlas. Thirdly, a score is computed for each processed landmark, thus allowing to assess the percentage of landmarks that lie within the \textit{confidence volumes}. The one feature of UPA-3DFA is that it can be applied to any kind of landmarks: either predicted by a 3DFA architecture/algorithm, or associated with an automatic or semi-automatic annotation process. }

\addnote[claims-3]{4}{The third contribution is a thorough experimental evaluation that uses two publicly available datasets and five 3DFA software packages \citep{3DFA1,3DFA2,3DFA3,3DFA4,3DFA5} associated with five peer-reviewed articles \citep{bulat2016two,feng2018joint,zhu2016face,tu2dasl19,guo2020towards}, respectively. The AFLW2000-3D dataset \citep{zhu2016face} and its semi-automatic annotations are used to compare UPA-3DFA with the supervised 
normalized mean error (NME). Correlation scores between UPA and NME are provided. This combined unsupervised-supervised analysis reveals the existence of annotation errors as well as a mechanism to disregard these errors. Altogether, this provides a novel methodological pipeline to evaluate the performance of 3DFA architectures as well as to analyse the quality of automatic and semi-automatic annotations.
}

\begin{figure}[t!]
\centering
\subfloat[3D landmarks predicted with \citep{bulat2016two}]{\includegraphics[width=.50\textwidth]{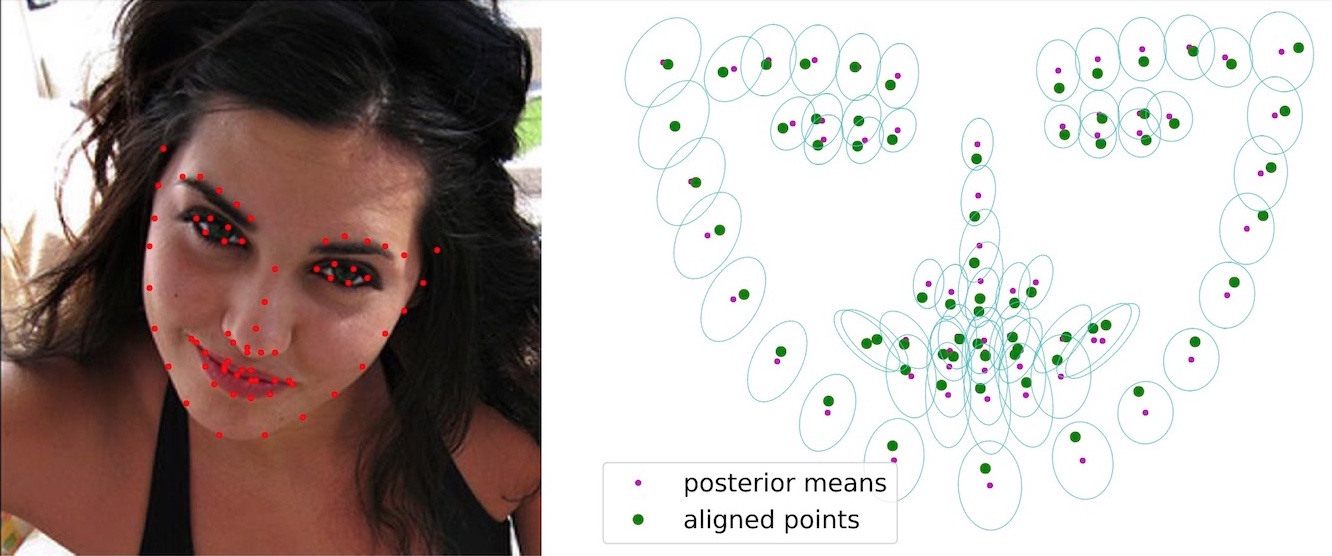}}\\
\subfloat[Semi-automatic annotated 3D landmarks from \citep{zhu2016face}]{\includegraphics[width=.50\textwidth]{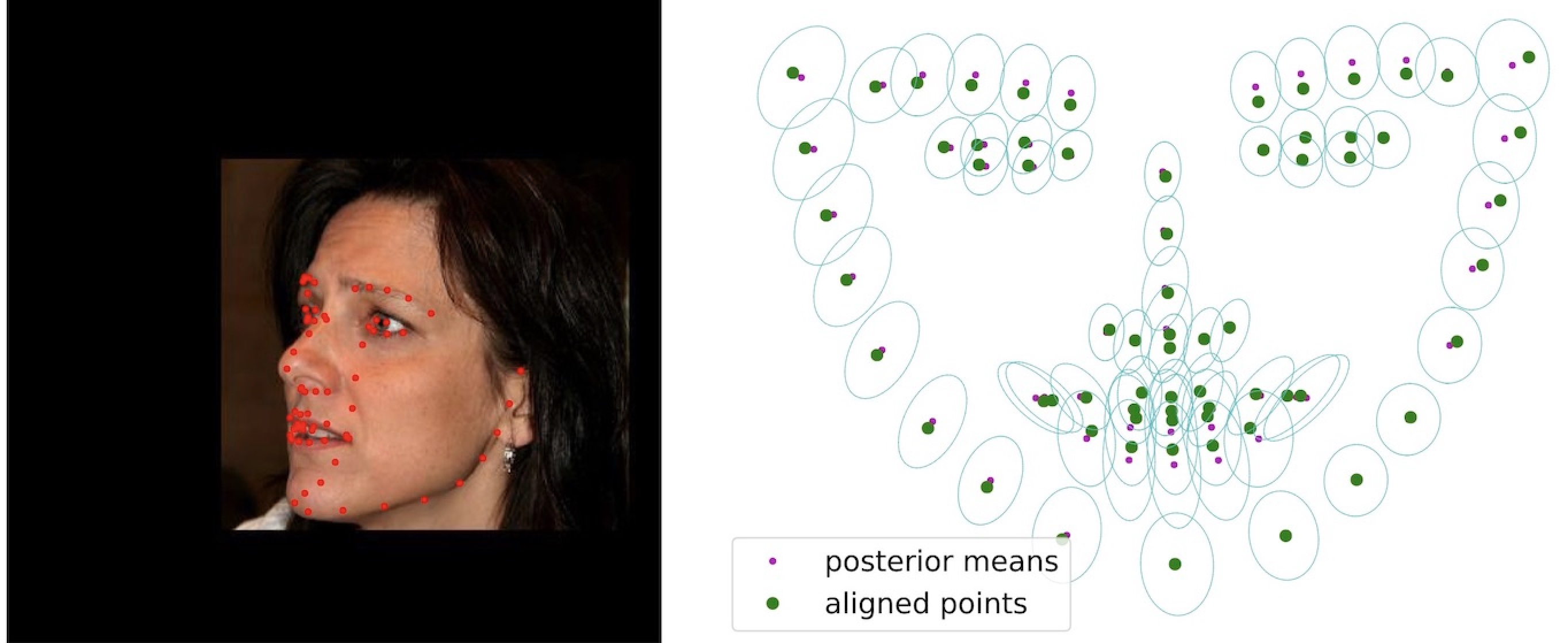}}
\caption{\label{fig:overview} Left: (a) landmarks predicted with the 3DFA method of \citep{bulat2016two}; (b) 3D landmarks obtained with the semi-automatic annotation method of \citep{zhu2016face}.
Right: The mapped landmarks (big green dots) are overlapped onto a frontal landmark model, or a shape atlas, composed of ellipsoids centered at mean landmark locations (small red dots). This enables one to verify whether tested landmarks (left) lie within ellipsoidal-shaped volumes of confidence (right). }
\end{figure}

The methodology is illustrated in Fig.~\ref{fig:overview} with two examples from the AFLW2000-3D dataset \citep{zhu2016face}. The statistical frontal landmark model (right) that is proposed consists of an ellipsoidal-shaped confidence volume centred at a posterior mean.
Fig.~\ref{fig:overview}(a): Landmarks extracted using \citep{bulat2016two} (left) are robustly mapped onto this frontal model (right). In this case, most of the landmarks lie inside their confidence volumes, thus assessing their correctness. Fig.~\ref{fig:overview}(b): 3D landmarks obtained with a semi-automatic annotation process \citep{zhu2016face} are robustly mapped onto the frontal model (right). Note that several annotated landmarks fall outside the confidence volumes. 

The remainder of this paper is organized as follows. Section~\ref{sec:related} reviews the related work. Section~\ref{sec:methodology} briefly reviews \ac{MLE} for rigid mapping and describes two robust methods. Section~\ref{sec:benchmark-robust} analyses the performance of the proposed rigid-mapping methods. Section~\ref{sec:statistical-face} describes a pipeline for building a statistical face model and an associated parametric confidence metric. Section~\ref{sec:experiments} presents extensive experimental results, and Section~\ref{sec:conclusions} draws some conclusions. The UPA-3DFA software is publicly available \citep{UPA-3DFA}.

\section{Related Work}
\label{sec:related}
\addnote[3dscans]{3}{
Until recently, 3D face landmarks were extracted from 3D scans, e.g. based on rotation-invariant curvature analysis \citep{szeptycki2009coarse}.}
In contrast, recently proposed 3DFA methods take as input 2D images and the underlying models lie at the crossroads of deformable shape models, model-based image analysis and DNNs. 
DNN-based 3DFA methods use a variety of architectures in order to learn a regression function, e.g. \citep{bulat2016two,feng2018joint,zhu2019face,tu2dasl19} as well as \citep{ning2020real,guo2020towards,hoang20213d}. Given this variety, it is difficult to directly compare them and assess their merits based on the analysis of the underlying DNN concepts and methodologies. Alternatively, 3DFA algorithm performance could be measured empirically, as is often the case in deep learning.

To date, there has been a handful of 3DFA benchmarks and challenges, \citep{phillips2005overview,jeni2016first,deng2019menpo,sanyal2019learning}. 
\addnote[scan-annotation]{3}{The 3D face scans of the FRGC dataset \citep{phillips2005overview} were used for benchmarking the rotation-invariant curvature-based method of  \citep{szeptycki2009coarse}. The authors manually annotated 12 facial landmarks. Note that this kind of annotation is possible because the 3D scans correspond to depth maps, and hence there is a depth value associated with each pixel location. Such a luxury is not possible with 2D images.}
In \citep{jeni2016first}, four datasets were specifically gathered, annotated and prepared, and two performance metrics were used for this challenge. The BU-4DFE \citep{yin126high} and BP-4D-Spontaneous \citep{zhang2014bp4d} datasets used a structured-light stereo sensor to capture textured 3D meshes of faces in controlled conditions and with various backgrounds. 2,295 meshes were selected from these datasets and manually annotated with 66 landmarks and with self-occlusion information. Then, 16,065 2D views were synthesized (seven views for each mesh) with yaw and pitch rotations ranging in the intervals $[-45^{\circ}, +45^{\circ}]$ and $[-30^{\circ}, +30^{\circ}]$, respectively.  Additionally, there were 7,000 frames from the Multi-PIE \citep{gross2010multi} and 541 frames from the Time-Sliced datasets, respectively. Both these datasets contain RGB images gathered with multiple cameras from different viewpoints but with no 3D information. Therefore, a 3D face model is extracted for each image, using the model-based multi-view structure-from-motion technique of \citep{jeni2017dense}. As above, each 3D face model was annotated with 66 landmarks and with self-occlusion information.


The Menpo challenge \citep{deng2019menpo} is based on a dataset of 12,000 face images. In order to obtain 2D and 3D ground-truth landmarks, an automatic annotation process is proposed, which fits a 3D face model to each 2D image (see above). This fitting is carried out via non-linear minimization over the shape parameters (identity and expression), the rigid parameters (rotation and translation of the 3D model with respect to the camera), and the camera parameters. 

The NoW benchmark \citep{sanyal2019learning} considers 3D reconstruction from a single monocular image of a face. 
The associated dataset contains 2,054 face images in frontal and profile views of 100 subjects and a 3D head scan for each subject. This dataset is similar in spirit with \citep{bagdanov2011florence}. While the images contain four categories (neutral, expression, occlusion, and selfie) the 3D scans correspond to neutral faces. Therefore, the challenge is the reconstruction of a neutral 3D face from a non-neutral 2D face, implying that the latter undergoes disentanglement. Moreover, since the predicted 3D face is a mesh and the ground-truth is a 3D scan (point cloud), that lie in different coordinate frames, a rigid alignment is needed.
The alignment method of \citep{sanyal2019learning}  minimizes a scan-to-mesh distance over the scale, rotation, and translation parameters. This leads to a non-linear optimization problem. In contrast, it is proposed a rigid point-to-point alignment technique that is robust with respect to errors and that preserves the facial expressions embedded in the 3D landmarks, rather than eliminating them. 

The evaluation metrics used in these benchmarks require either annotated 3D landmarks or 3D scans. As already argued, manual annotation is infeasible. Automatic annotation is based on complex non-linear minimization methods that are prone to errors and may not be reliable in the presence of profile views, of extreme expressions, and of occlusions. Localization noise is inherent. Moreover, these evaluation metrics are limited in scope since they cannot distinguish between landmark localization noise (inlying data) and large localization errors (outlying data).

In contrast, the proposed methodology doesn't require annotations of any kind. Robust rigid alignment (analyzed in detail below) is used to build a frontal landmark model, i.e. a shape atlas, in a completely unsupervised way. A statistical characterization of each landmark is provided by measuring the discrepancy between the predicted landmarks and the corresponding model landmarks. 
\addnote[expression]{2}{
This is particularly useful to check whether the predicted 3D landmarks could be used any further, e.g. for facial expression recognition,  lip reading, or head-pose estimation.
In addition, the proposed analysis may well be used to remove badly located landmarks from an automatically annotated dataset.}


\addnote[robust-gum]{3}{A fundamental building block of the proposed method is a robust rigid transformation estimator. Robustness refers to the capacity of an estimator to be unaffected by large errors, i.e. outliers. For that purpose, the choice of a probability distribution function is crucial. We opt for two choices: (i)~a mixture model formed by a Gaussian component and a uniform component (GUM), and  (ii)~the generalized Student's t-distribution (GStudent) \citep{McLachlanPeel2000b,sun2010robust,forbes2014new,chamroukhi2017skew}. Robust mixture models that use a Gaussian mixture with an additional uniform component have been used for several decades in model based clustering, in order to downgrade the influence of outliers \citep{banfield1993model}. They have also been used for robust factorization \citep{zaharescu2009robust}, for point-set registration \citep{myronenko2010point} and, more recently, for robust deep regression \citep{lathuiliere2018deepgum}. The EM algorithm proposed in \citep{myronenko2010point} alternates between point registration (matching) and rigid alignment, while the uniform component of the mixture is used to disregard points in one set that don't have a match in the other set. Our use of GUM is different. Since the points (landmarks) are already registered, the residuals present in \eqref{eq:registration-general} (please refer to the next Section) are drawn from a GUM distribution in order to model shape-, deformation- and localization errors.
}

\addnote[gum-student]{3}{
GUM and GStudent treat the above errors quite differently. GUM evaluates the posterior probability of a data point to be either an inlier or an outlier, i.e. \eqref{eq:post-GUM} and \eqref{eq:posterior-gum}. GStudent evaluates a weight $w$ for each data point. The weights are treated as random variables drawn from a gamma distribution, i.e. \eqref{eq:generalized-student} and they can be interpreted as precisions (the inverse of the variance): higher is a weight, more reliable is the corresponding data point. While the GUM posteriors can only vary in the range $[0;1]$, the weight realizations vary from zero to a very large positive value. Both GUM and GStudent provide a statistically well-founded mechanism to associate a figure of merit to each data point and hence to downgrade the influence of large localization errors. This is of paramount importance, not only to properly estimate rigid parameters from 3D landmarks (Section~\ref{sec:methodology}), but also to build a robust confidence test (Section~\ref{sec:statistical-face}).}



%
%
%
%

\section{Robust Rigid Mapping}
\label{sec:methodology}

Let us consider the mapping between two sets of 3D facial landmarks, a predicted set, $\xvect_{1:N} = ( \xvect_1 \dots \xvect_N) \in \mathbb{R}^{3\times N}$, and a model set, $\yvect_{1:N} = (\yvect_1 \dots \yvect_N) \in \mathbb{R}^{3\times N}$. 
\addnote[mapping-model]{1}{The predicted set corresponds to a face with arbitrary and unknown pose, identity, expression and occlusion.
Without loss of generality, the model set corresponds to a frontally viewed neutral face.
The mapping writes:}
\begin{equation}
\label{eq:registration-general}
 \yvect_{n}  =  s \Rmat \xvect_{n} +  \tvect + \rvect_{n}, \forall n \in \{1,\dots, N\},
\end{equation}
where the rigid transformation is parameterized by a scale factor $s\in\mathbb{R}$, a rotation matrix $\Rmat\in SO(3) \subset \mathbb{R}^{3\times 3}$ and a translation vector $\tvect\in\mathbb{R}^{3}$, while the non-rigid deformations and possible errors are modeled by the residuals $\rvect_{1:N}$. 
Let's cast the rigid-mapping estimation problem into the framework of maximum-likelihood estimation (MLE) with a robust pdf.
It is assumed that the residuals $\rvect_{1:N}$ are independent and identically distributed (i.i.d). Then, the problem of estimating the rigid transformation could be solved via MLE:
\begin{equation}
\label{eq:log-likelihood}
\mathcal{L}(\thetavect | (\xvect, \yvect)_{1:N}) = - \sum_{n=1}^{N} \log P (\rvect_n; \thetavect),
\end{equation}
where $P (\rvect; \thetavect)$ is the pdf of $\rvect$ parameterized by $\thetavect$. 
\subsection{Gaussian Model}
\label{sec:gaussian}
The simplest statistical model is to assume that the residuals follow a zero-centered Gaussian distribution with covariance matrix $\Sigmamat\in\mathbb{R}^{3\times 3}$, namely $P (\rvect; \thetavect) = \mathcal{N}(\rvect; \zerovect, \Sigmamat)$. Equation \eqref{eq:log-likelihood} yields:
\begin{equation}
\label{eq:log-likelihood-gaussian}
\mathcal{L}(\thetavect |  (\xvect, \yvect)_{1:N}) =  \sum_{n=1}^{N} \big( \| \yvect_n - s \Rmat \xvect_n - \tvect \|^2_{\Sigmamat} + \log |\Sigmamat| \big),
\end{equation}
where
$\| \avect \|_{\Sigmamat}^2= \avect\tp \Sigmamat\inverse\avect$ is the squared Mahalanobis norm of $\avect\in\mathbb{R}^3$ and $|\cdot |$ is the determinant operator.
The minimization of \eqref{eq:log-likelihood-gaussian} over $\tvect$ yields:
\begin{equation}
\label{eq:optimal-translation}
\tvect^{\star} = \ovec{y} - s^{\star} \Rmat^{\star}  \ovec{x},
\end{equation}
where $\pvect^{\star}$ is the optimal value of a parameter $\pvect$, and with:
\begin{align}
\label{eq:xy-scaled}
\ovec{x} = 1/N\sum_{n=1}^{N} \xvect_n,  \quad
\ovec{y} = 1/N\sum_{n=1}^{N} \yvect_n.
\end{align}
By substituting \eqref{eq:optimal-translation} into \eqref{eq:log-likelihood-gaussian} and by using centered coordinates, i.e.    
$\xvect^{\prime}_n = \xvect_n - \ovec{x}$, $\yvect^{\prime}_n = \yvect_n - \ovec{y}$,
one obtains:
\begin{equation}
\label{eq:optimization-rotation-scale}
\mathcal{L}(\thetavect |  (\xvect^{\prime}, \yvect^{\prime})_{1:N}) =  \sum_{n=1}^{N} \big(  \| \yvect^{\prime}_n - s \Rmat \xvect^{\prime}_n \|^2_{\Sigmamat} + \log |\Sigmamat| \big).
\end{equation}
\addnote[iso-cov]{3}{
Standard approaches assume an isotropic covariance,
$\Sigmamat=\sigma \Imat_3$, yielding a closed-form solution, e.g. \citep{horn1987closed,Umeyama91} and \ref{app:rotation-unitquaternion}.
Indeed, one may easily verify that in this particular case
the $s^2 \xvect_n^{\prime} \Rmat \Sigmamat\inverse \Rmat\tp \xvect_n^{\prime\top}$ term in the development of $ \| \yvect^{\prime}_n - s \Rmat \xvect^{\prime}_n \|^2_{\Sigmamat} $  that is present in \eqref{eq:optimization-rotation-scale} is equal to
$ s \sigma\inverse  \xvect_n^{\prime}   \xvect_n^{\prime\top}$.
Nevertheless, the isotropic-co\-va\-riance assumption is barely valid in practice. 
In the case of a full co\-va\-riance, the minimization of \eqref{eq:optimization-rotation-scale} with respect to the rotation matrix yields:
\begin{equation}
\label{eq:optimization-rotation}
\Rmat^{\star}  = \argmin{_{\Rmat}} \trace \big( \Sigmamat\inverse ( s^2 \Rmat \Amat \Rmat\tp - 2 s \Rmat \Bmat )\big),
\end{equation}
where $\trace (\cdot)$ is the trace operator and with the notations
 $\Amat= \sum_{n=1}^{N} \xvect^{\prime}_n \xvect_n^{\prime\top}$,  $\Bmat= \sum_{n=1}^{N} \xvect^{\prime}_n \yvect_n^{\prime\top}$. 
A rotation matrix must satisfy $\Rmat\Rmat\tp=\Imat_3$ and $|\Rmat|=+1$. This yields a constrained non-linear optimization problem. An elegant formulation consists of parameterizing the rotation with a unit quaternion, thus reducing the number of parameters from 9 to 4, and the number of constraints from 7 to 1.
 Let
 $\Rmat(\qvect)$, where the parameter vector $\qvect$ is a unit quaternion, i.e. \ref{app:rotation-unitquaternion}. The constrained non-linear optimization problem, i.e. \ref{app:implementation}, writes:
\begin{equation}
\label{eq:optimization-quaternion}
\begin{cases}
\min_{\qvect}  &
\trace \big( \Sigmamat\inverse ( s^2 \Rmat (\qvect) \Amat \Rmat (\qvect)\tp - 2 s \Rmat (\qvect)  \Bmat )\big) \\
\textrm{s.t.} & \qvect\tp\qvect = 1
\end{cases}
\end{equation}
}
\subsection{Gaussian-uniform Mixture Model}
\label{sec:gum}
Unfortunately, the above statistical model does not behave well in the presence of large residuals, or outliers.
A discrete hidden random variable $Z_n$ is associated with each residual $\rvect_n$, and let $z$ be a realization of $Z$. $\rvect$ is drawn either from a zero-centered Gaussian distribution or from a multivariate uniform distribution:
\begin{equation}
\label{eq:post-GUM}
P(\rvect | Z=z) = 
\begin{cases}
\mathcal{N} (\rvect; \zerovect, \Sigmamat) & \mathrm{if} \; z=\mathrm{inlier} \\
\mathcal{U}(\rvect; 0, \gamma) & \mathrm{if} \; z=\mathrm{outlier},
\end{cases}
\end{equation}
where $\gamma$ is the volume of the distribution. This yields a two-component mixture model, an inlier component with prior $p$, and an outlier component with prior $1-p$. Formally, this leads to solving the problem via expectation-maximization (EM) which alternates betwen (i)~evaluating the posterior probabilities of the residuals, and (ii)~minimizing the \textit{expected complete-data negative log-likeli\-hood}, $\mathrm{E}_Z [ - \log P( \rvect_{1:N}, Z_{1:N} | \rvect_{1:N} ; \thetavect)]$,
where the expectation is taken over the realizations of $Z$, with $\thetavect = \{s, \Rmat, p, \Sigmamat\}$.\footnote{Note that the translation vector $\tvect$ is evaluated with \eqref{eq:optimal-translation}.} This yields the minimization of:
\begin{equation}
\label{eq:expectation-rotation-scale}
\mathcal{E}(\thetavect |  (\xvect^{\prime} , \yvect^{\prime} )_{1:N}) =  \sum_{n=1}^{N} \alpha_n \big(  \| \yvect^{\prime}_n - s \Rmat \xvect^{\prime}_n \|^2_{\Sigmamat} + \log |\Sigmamat| \big)
\end{equation}
where the inlier posterior $\alpha_n = P(Z=\mathrm{inlier}| \rvect_n)$, is :
\begin{equation}
\label{eq:posterior-gum}
\alpha_n = \frac{p \: \mathcal{N} (\rvect_n; \zerovect, \Sigmamat)}
{p \: \mathcal{N} (\rvect_n; \zerovect, \Sigmamat) + (1-p) \: \gamma\inverse},
\end{equation}
and the outlier posterior is $1-\alpha_n$. 
The presence of $\alpha_{1:N}$ in \eqref{eq:expectation-rotation-scale} replaces \eqref{eq:xy-scaled} with:
\begin{align}
\label{eq:xy-scaled-weighted}
\ovec{x} = \sum_{n=1}^{N}\alpha_n  \xvect_n / \sum_{n=1}^{N}\alpha_n ,  \quad
\ovec{y} = \sum_{n=1}^{N}\alpha_n  \yvect_n / \sum_{n=1}^{N}\alpha_n ,
\end{align}
as well as $\Amat$ and $\Bmat$ from \eqref{eq:optimization-quaternion} with 
\begin{equation}
\label{eq:AB-weighted}
\Amat= \sum_{n=1}^{N} \alpha_n \xvect^{\prime}_n \xvect_n^{\prime\top}, \quad \Bmat= \sum_{n=1}^{N} \alpha_n \xvect^{\prime}_n \yvect_n^{\prime\top}.
\end{equation}
Hence, \eqref{eq:optimization-quaternion} can be used to estimate the optimal rotation while the optimal scale is estimated with:
\begin{equation}
\label{eq:optimal-scale-weighted}
s^{\star} = \left(
\frac
{\sum_{n=1}^{N} \alpha_n \yvect_n^{\prime\top} \Sigmamat\inverse \yvect^{\prime}_n }
{\sum_{n=1}^{N} \alpha_n (\Rmat^{\star} \xvect^{\prime}_n){^{\top}} \Sigmamat\inverse  \Rmat^{\star} \xvect^{\prime}_n }
\right)^{1/2}.
\end{equation}
The prior $p$ and covariance $\Sigmamat$ are estimated with:
\begin{align}
\label{eq:prior-gum}
p &= \frac{1}{N} \sum_{n=1}^{N} \alpha_n,  \\
\label{eq:covariance-empirical-gum}
\Sigmamat^{\star} &= \frac{ \sum_{n=1}^{N} \alpha_n (\yvect^{\prime}_n - s^{\star} \Rmat^{\star} \xvect^{\prime}_n) (\yvect^{\prime}_n - s^{\star} \Rmat^{\star} \xvect^{\prime}_n)\tp}
{\sum_{n=1}^{N} \alpha_n}.
\end{align}
This model is referred to as the \textit{Gaussian-uniform mixture} (GUM) and the associated EM is summarized in Algorithm~\ref{algo:em-robfa}.

\begin{algorithm}[t!]
 \caption{\label{algo:em-robfa} GUM Expectation-Maximization.}
 \KwData{Centered point coordinates, i.e. \eqref{eq:xy-scaled}. Normalization parameter $\gamma$ \;}

 \textbf{Initialization of $\thetavect^{\mathrm{old}}$}: Use the closed-form solution \citep{horn1987closed} to evaluate $s^{\mathrm{old}}$ and $\Rmat^{\mathrm{old}}$ and then use  these parameter values to evaluate $\Sigmamat^{\mathrm{old}}$ and set $p^{\mathrm{old}}=0.8$\;
 
 \While{$\|\thetavect^{\mathrm{new}} - \thetavect^{\mathrm{old}}\|>\epsilon$}{
 \textbf{E-step}: Evaluate the posteriors $\alphavect_{1:N}$ using \eqref{eq:posterior-gum} with $\thetavect^{\mathrm{old}}$\;
 
 Update the centered coordinates using \eqref{eq:xy-scaled-weighted} \;

 \textbf{M-scale-step}:   Evaluate $s^{\mathrm{new}}$ using \eqref{eq:optimal-scale-weighted}\;
 
 \textbf{M-rotation-step}: Estimate $\Rmat^{\mathrm{new}}$  via constrained non-linear optimization of \eqref{eq:optimization-quaternion} using \eqref{eq:AB-weighted} \;
 
 \textbf{M-covariance-step}:  Evaluate $\Sigmamat^{\mathrm{new}}$ using \eqref{eq:covariance-empirical-gum}\;
 
  \textbf{M-prior-step}:   Evaluate $p^{\mathrm{new}}$ using \eqref{eq:prior-gum}\;
  
  $\thetavect^{\mathrm{old}} \leftarrow \thetavect^{\mathrm{new}}$\;
 }
 
 
  \KwResult{Estimated scale $s^{\star}$, rotation $\Rmat^{\star}$, translation $\tvect^{\star}$ \eqref{eq:optimal-translation}, prior $p^{\star}$, covariance $\Sigmamat^{\star}$, and posterior probabilities of landmarks $\alphavect_{1:N}$.}

\end{algorithm}

\subsection{Generalized Student Model}
\label{sub:student-distribution}
Another way to enforce robustness is to use the \textit{generalized Student's t-distribution}, also known as the Pearson type~VII distribution \citep{sun2010robust}:
\begin{align}
\label{eq:generalized-student}
P(\rvect; \Sigmamat, & \mu, \nu)   = \int_{0}^{\infty} \mathcal{N} ( \rvect; 0, w \inverse \Sigmamat) \mathcal{G}(w, \mu, \nu) dw \nonumber \\
&=
\frac{\Gamma(\mu + \frac{3}{2})}{| \Sigmamat |^{\frac{1}{2}} \Gamma(\mu) (2\pi \nu)^{\frac{3}{2}}}
\left(
1+ \frac{\| \rvect \|^2_{\mathcal{M}}} {2\nu}
\right)_{,}^{-\left(\mu + \frac{3}{2}\right)}
\end{align}
where $\Gamma(\cdot)$ is the gamma function.
Note that the \textit{latent} weight variables $w_{1:N}$ are drawn from a prior gamma distribution with parameters $\mu$ and $\nu$. 
In \eqref{eq:generalized-student} $\nu$ and $\Sigmamat$ appear only through their product, which means that an additional constraint is required to make the parameterization unique. 
Let $\nu=1$ as in \citep{forbes2014new}.
The posterior distribution of $w_n$ is also a gamma distribution, namely the \textit{posterior gamma distribution}:
\begin{align}
\label{eq:weight-gamma-posterior}
P(w_n | \rvect_n; \Sigmamat, \mu, \nu) &= \mathcal{N} ( \rvect_n; 0, w_n \inverse \Sigmamat) \mathcal{G}(w_n, \mu, \nu) \nonumber \\
& = \mathcal{G} (w_n; a, b_n),
\end{align}
with parameters:
\begin{align}
\label{eq:gamma-posterior-ab}
a  = \mu + 3/2, \quad
b_n = 1 + \| \rvect_n \| ^2_{\Sigmamat}/2.
\end{align}
The posterior mean of the weight variable is:
\begin{equation}
\label{eq:weight-expectation}
\overline{w}_n = \mathrm{E}[w_n | \rvect_n] = a / b_n.
\end{equation}
As above, one needs to minimize the expected complete-data negative log-likeli\-hood, $\mathrm{E}_W [ - \log P( \rvect_{1:N}, W_{1:N} | \rvect_{1:N} ; \thetavect)]$ with $\thetavect = \{s, \Rmat, \Sigmamat, \mu\}$, yielding the minimization:
\begin{equation}
\label{eq:expectation-student}
\mathcal{Q}(\thetavect |  (\xvect^{\prime} , \yvect^{\prime} )_{1:N}) =  \sum_{n=1}^{N} \big( \overline{w}_n \| \yvect^{\prime}_n - s \Rmat \xvect^{\prime}_n \|^2_{\Sigmamat} + \log |\Sigmamat| \big),
\end{equation}
thus replacing $\alpha_{1:N}$ with $w_{1:N}$ in \eqref{eq:xy-scaled-weighted} and \eqref{eq:AB-weighted} to estimate the optimal rotation \eqref{eq:optimization-quaternion} and scale \eqref{eq:optimal-scale-weighted}.
The covariance matrix is estimated with:
\begin{equation}
\label{eq:covariance-student}
\Sigmamat = \sum_{n=1}^{N} \overline{w}_n (\yvect^{\prime}_n - s \Rmat \xvect^{\prime}_n) (\yvect^{\prime}_n - s \Rmat \xvect^{\prime}_n)\tp / N
\end{equation}
The parameter $\mu$ is updated by solving the following equation, where $\Psi(\cdot)$ is the digamma function:
\begin{equation}
\label{eq:mu-digamma}
\mu = \Psi^{-1}\left( \Psi(a) -\frac{1}{n} \sum_{n=1}^{N} \log b_n  \right).
\end{equation}
This model is referred to as the \textit{generalized Student} (GStudent) and the associated EM algorithm is summarized in Algorithm~\ref{algo:em-student}.

\begin{algorithm}[t!]
 \caption{\label{algo:em-student} GStudent Expectation-Maximization.}

 \KwData{Centered point coordinates, i.e. \eqref{eq:xy-scaled}. \;}

 \textbf{Initialization of $\thetavect^{\mathrm{old}}$}: Use the closed-form solution \citep{horn1987closed} to evaluate $s^{\mathrm{old}}$ and $\Rmat^{\mathrm{old}}$;  evaluate $\Sigmamat^{\mathrm{old}}$. Provide $\mu^{\textrm{old}}$ \;
 
 \While{$\|\thetavect^{\mathrm{new}} - \thetavect^{\mathrm{old}}\|>\epsilon$}{
 \textbf{E-step}: evaluate $a^{\mathrm{new}}$ and $\bvect_{1:N}^{\mathrm{new}}$ using \eqref{eq:gamma-posterior-ab} with $\thetavect^{\mathrm{old}}$, then evaluate $\overline{\wvect}_{1:N}^{\mathrm{new}}$ using \eqref{eq:weight-expectation} \;
 
 Update the centered coordinates using \eqref{eq:xy-scaled-weighted}, where $\alpha_{1:N}$ are replaced with $\overline{w}_{1:N}$ \;

 \textbf{M-scale-step}:   Evaluate $s^{\mathrm{new}}$ using \eqref{eq:optimal-scale-weighted}\;
 
 \textbf{M-rotation-step}: Estimate $\Rmat^{\mathrm{new}}$  with \eqref{eq:optimization-quaternion}, \eqref{eq:AB-weighted} \;
 
 \textbf{M-covariance-step}:  Evaluate $\Sigmamat^{\mathrm{new}}$ using \eqref{eq:covariance-student} \;
 
  \textbf{M-mu-step}:   Evaluate $\mu^{\mathrm{new}}$ using \eqref{eq:mu-digamma} \;
  
  $\thetavect^{\mathrm{old}} \leftarrow \thetavect^{\mathrm{new}}$\;
 }
 
 
  \KwResult{Estimated scale $s^{\star}$, rotation $\Rmat^{\star}$, translation $\tvect^{\star}$ \eqref{eq:optimal-translation}, covariance $\Sigmamat^{\star}$, and landmark weights  $w_{1:N}$.}

\end{algorithm}

\section{Analyzing the Robustness of Rigid Mapping}
\label{sec:benchmark-robust}
\begin{figure*}[t!]
\centering 
\subfloat[Rotation error.]{\includegraphics[width=.45\textwidth,trim={0.cm, 0.cm, 0cm, 0cm},clip]{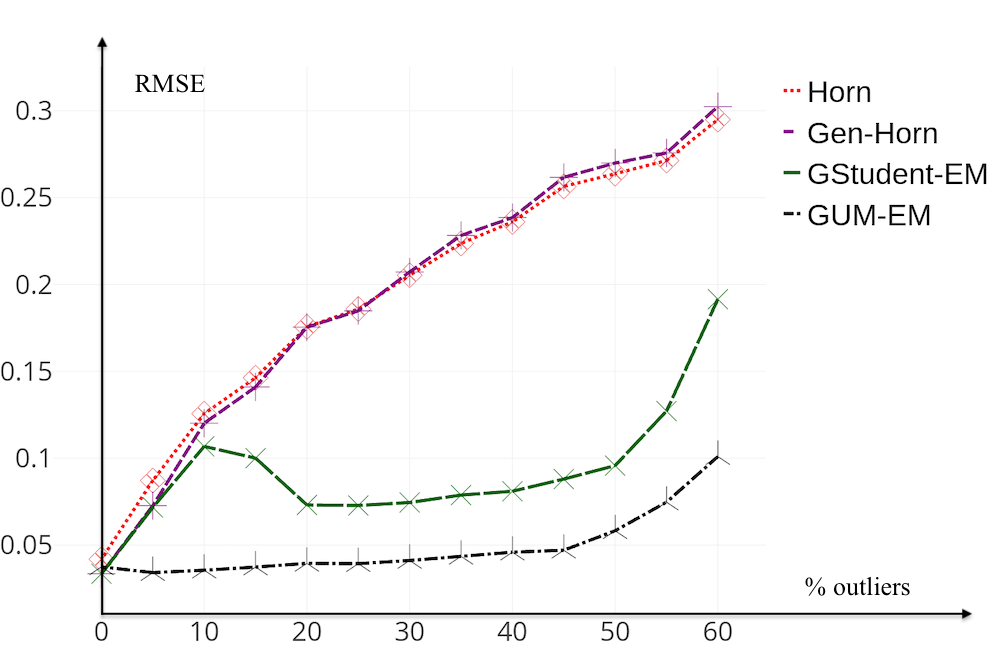}}
\subfloat[Translation error.]{\includegraphics[width=.45\textwidth,trim={0.cm, 0.cm, 0cm, 0cm},clip]{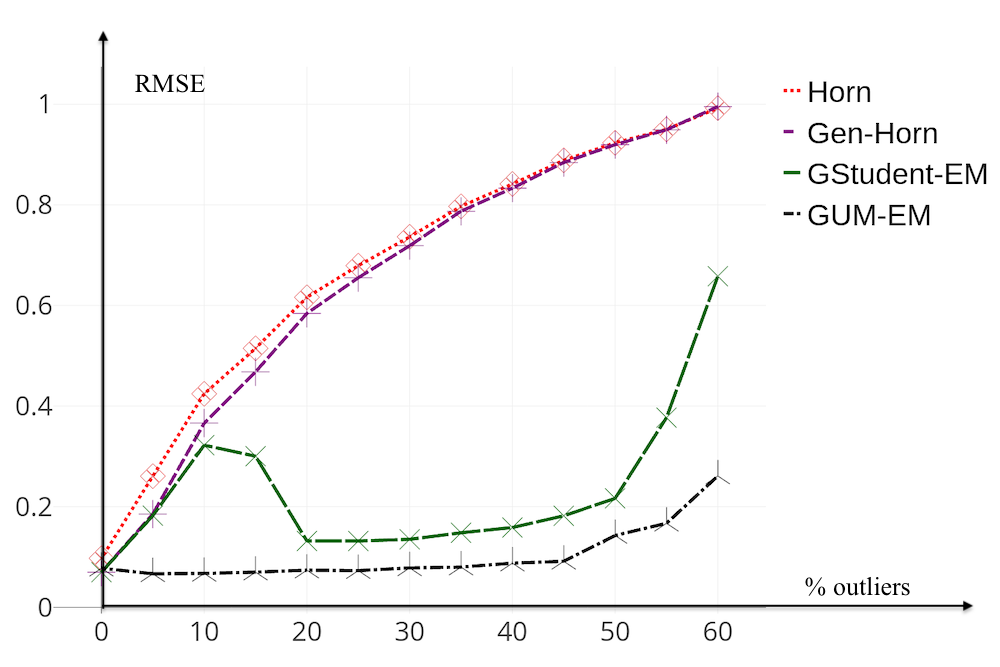}}
\caption{\label{fig:GUaniso_incRatio_rmse} RMSE as a function of the percentage of outliers ($0\%$ to $60\%$): inliers are affected by Gaussian noise with $\lambda=0.0025$ while outliers are affected by uniform noise with amplitude $a=1.5$.}
\end{figure*}


\begin{figure*}[t!]
    \centering
    \subfloat[Rotation error.]{\includegraphics[width=.45\textwidth,trim={0.cm, 0.cm, 0cm, 0cm},clip]{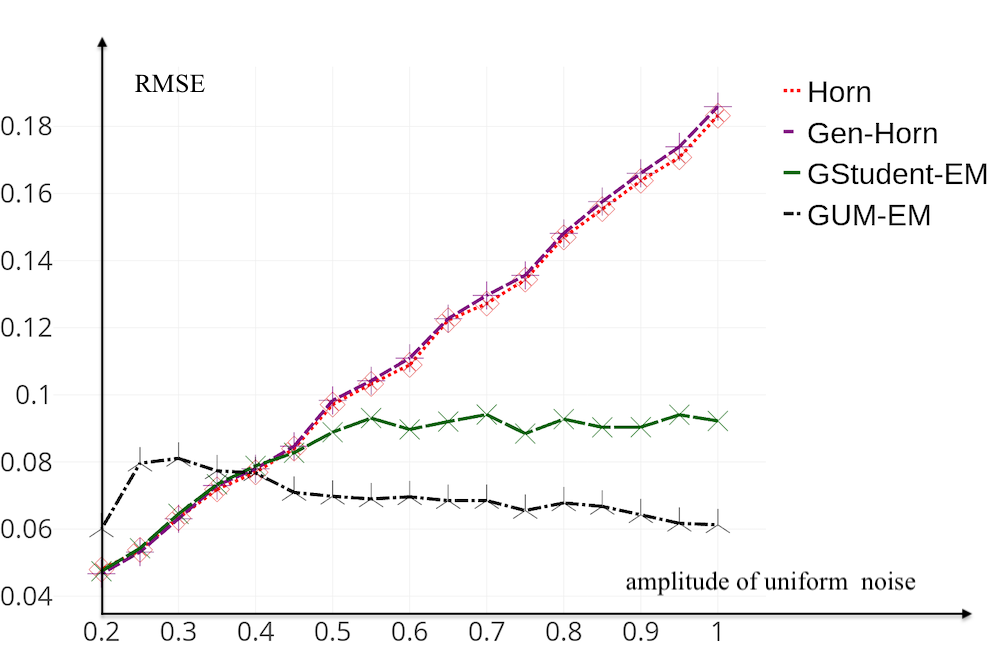}}
    \subfloat[Translation error.]{\includegraphics[width=.45\textwidth,trim={0.cm, 0.cm, 0cm, 0cm},clip]{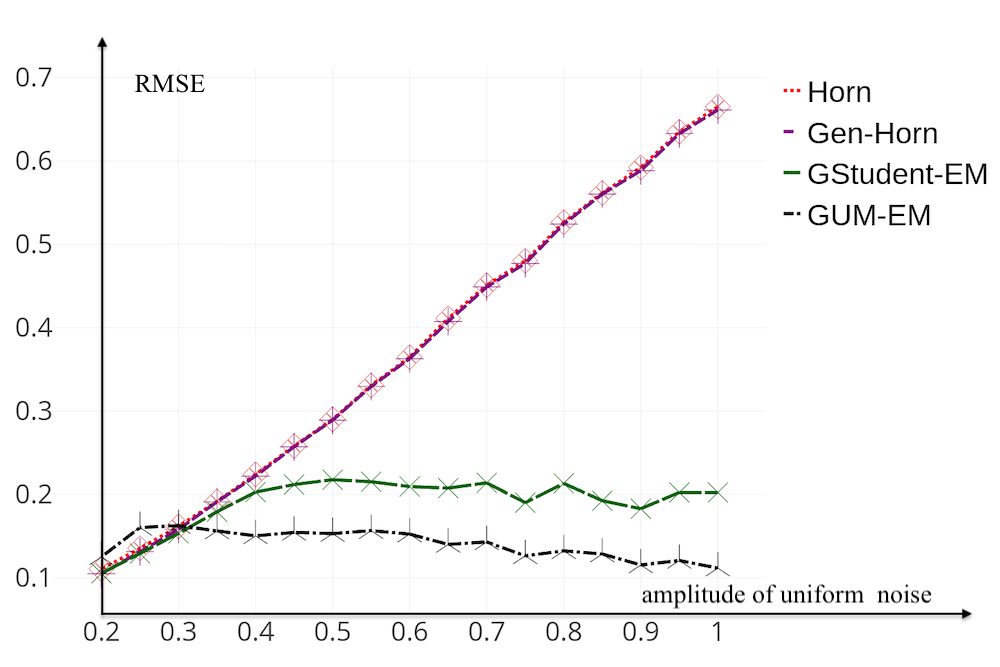}}
\caption{\label{fig:GUaniso_incAmp_rmse} RMSE as a function of uniform noise affecting a fixed number of outliers: inliers (50\%) are affected by Gaussian noise with $\lambda=0.0025$ while outliers (50\%) are affected by uniform noise of increasing amplitude $a\in [0.2,1.0]$.}
\end{figure*}

\begin{figure*}[t!]
    \centering
    \subfloat[Rotation error.]{\includegraphics[width=.45\textwidth,trim={0.cm, 0.cm, 0cm, 0cm},clip]{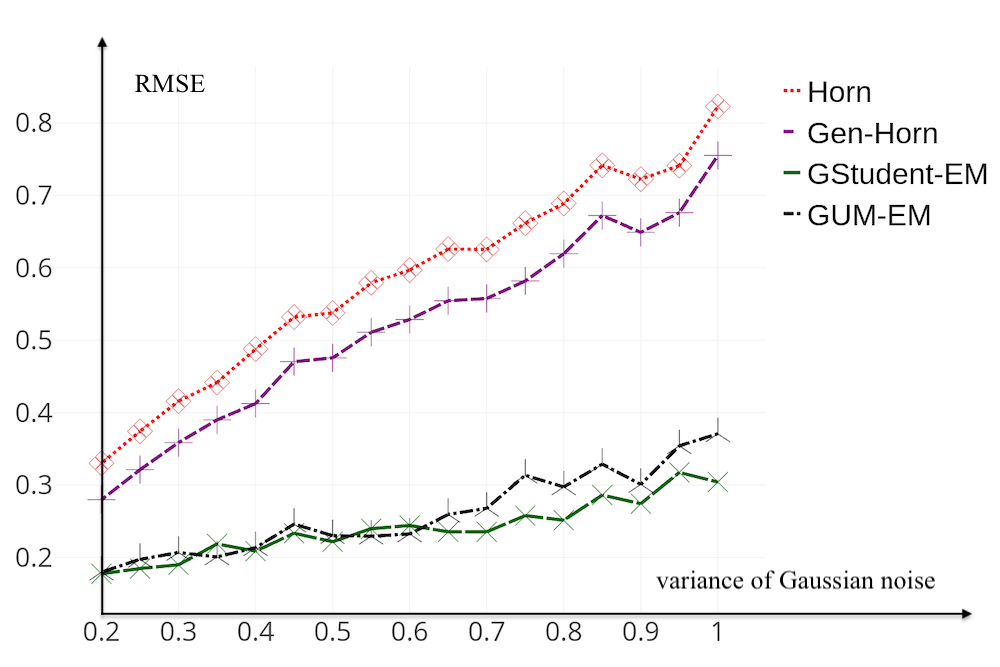}}
    \subfloat[Translation error.]{\includegraphics[width=.45\textwidth,trim={0.cm, 0.cm, 0cm, 0cm},clip]{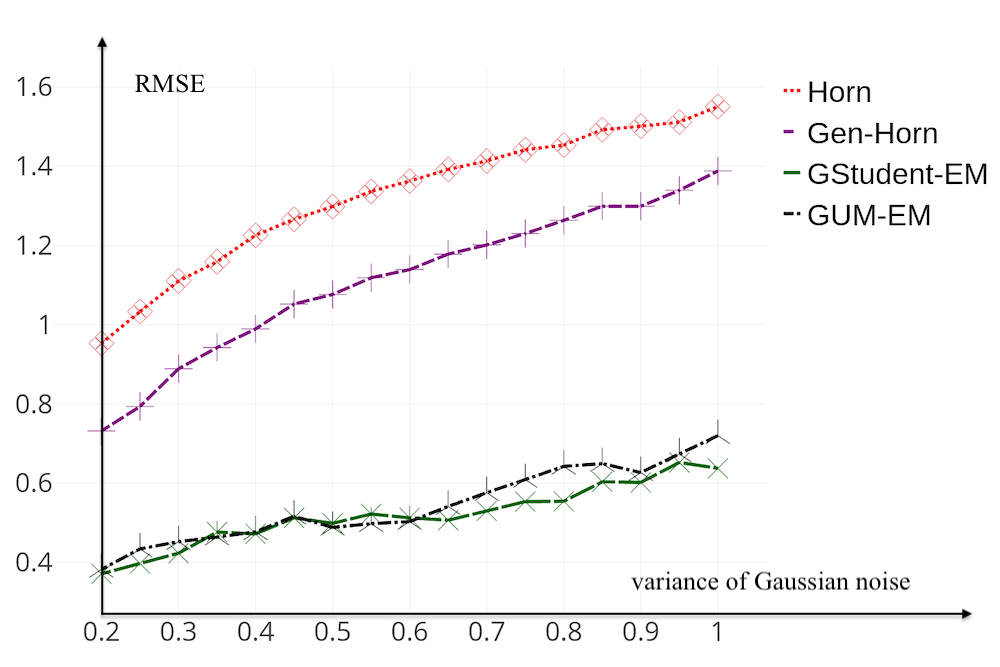}}
\caption{\label{fig:GGaniso_incAmp_rmse} RMSE as a function of Gaussian noise affecting the outliers: inliers (50\%) are affected by Gaussian noise with $\lambda=0.0025$ while outliers (50\%) are affected by Gaussian noise with $\lambda\in [0.2,1.0]$.}

\end{figure*}

In order to quantify the performance of the proposed robust rigid-mapping algorithms, An experimental protocol is devised on the following grounds. As above:
\begin{equation}
 \yvect_{n}^m(b)  =  s^m \Rmat^m \xvect_{n} +  \tvect^m + \rvect_{n}^m(b), \forall n\in \{1, \dots, N\},
\end{equation}
where $b>0$ is a scalar that controls the level of noise and $m$ is the trial index. 
As described in detail below, the noise level, $b$ can be the total variance of Gaussian anisotropic noise, or the volume of uniformly distributed noise.
The image coordinates are normalized such that $\forall n, \xvect_n \in [0,1]^3$. For each noise level, $M$ trials are randomly generated, namely $M$ rigid mappings and $M$ sets of $N$ residuals $\rvect_{1:N} = \{\rvect_{n}\}_{n=1}^{n=N}$. 
For each trial $m$ the rigid mapping  parameters are estimated, $s^m$, $\Rmat^m$, $\tvect^m$, and the \textit{root mean square error}, (RMSE) between these estimated parameters and the ground-truth parameters $\tilde{s}^m$, $\tilde{\Rmat}^m$,  $\tilde{\tvect}^m$, is estimated, namely:
\begin{align}
\label{eq:rmse-scale}
E_s &=  \Big( 1/M \textstyle {\sum_{m=1}^M} | s^m - \tilde{s}^m |^2 \Big)^{1/2}, \\
\label{eq:rmse-translation}
E_{\tvect} &= \Big( 1/M \textstyle{\sum_{m=1}^M} \|  \tvect^m -  \tilde{\tvect}^m \|^2 \Big)^{1/2} , \\
\label{eq:rmse-rotation}
E_{\Rmat} &=  \Big( 1/M \textstyle{\sum_{m=1}^M} \| \Rmat^m - \tilde{\Rmat}^m \|^2 \Big)^{1/2},
\end{align}
The ground-truth rigid-mapping parameters are generated in the following way.  For each trial $m$, the scale and the translation vector are generated from uniform distributions, namely $s^m \sim \mathcal{U}(0.5,2)$ and $\tvect^m \sim \mathcal{U}(0.5,5)^3$. The rotation matrix is parameterized by the pan, tilt and yaw angles, namely, $\Rmat  = \Rmat_\gamma \Rmat_\phi \Rmat_\psi$.
A rotation matrix is obtained by randomly generating the pan, tilt and yaw angles, $\gamma^m, \phi^m$, $\psi^m$, from a uniform distribution, $\mathcal{U} (-90^\circ, +90^\circ)$.

In order to generate residuals, $\rvect_{1:N}$, Gaussian and uniform noise are simulated, namely $ \rvect \sim \mathcal{N} (\zerovect, \Sigmamat)$ and $ \rvect \sim \mathcal{U} (-a/2, a/2)^3$.
In the  Gaussian case, a covariance matrix must be randomly generated for each trial. This is done in the following way. Let $\Sigmamat=\Qmat\Lambdamat\Qmat{\tp}$, with $\Qmat\in O(3)$ and with $\Lambdamat = \diag(\lambda_1, \lambda_2,\lambda_3)$. 
Let $\lambda=\lambda_1+\lambda_2+\lambda_3$, i.e. the total variance.
A sample covariance $\Sigmamat$ is simulated by randomly generating an orthogonal matrix $\Qmat$ as well as three eigenvalues from a uniform distribution, $\mathcal{U} (0,1)$.

The following rigid mapping models and associated algorithms were tested:
\begin{itemize}
\item \textit{Horn}: Gaussian distribution with isotropic covariance, \citep{horn1987closed} and \ref{app:rotation-unitquaternion};
\item \textit{Gen-Horn}: Gaussian distribution with anisotropic covariance, Section~\ref{sec:gaussian};
\item \textit{GUM-EM}: Gaussian-uniform mixture distribution, Algorithm~\ref{algo:em-robfa}, and
\item \textit{GStudent-EM}: Generalized Student's t-distribution, Algorithm~\ref{algo:em-student}.
\end{itemize}

The experiments were conducted in the following way. For each noise level, $M=500$ trials were simulated and the RMSEs were computed, namely eqs.~\eqref{eq:rmse-scale}, \eqref{eq:rmse-translation}, and \eqref{eq:rmse-rotation}. For each trial $m$ the landmarks were split into an inlier set and an outlier set and the $N=68$ landmarks are randomly assigned to one of these sets. The first experiment determines the percentage of outliers that can be handled by the robust algorithms, Figure~\ref{fig:GUaniso_incRatio_rmse}. For this purpose, the percentage of outliers is increased from 10\% to 60\%. The inlier noise is drawn from an anisotropic Gaussian distribution with a total variance $\lambda=0.0025$. The outlier noise is drawn from a uniform distribution with amplitude $a=1.5$ (remember that the landmark coordinates are normalized to lie in the interval $[0,1]$). The curves plotted in Figure~\ref{fig:GUaniso_incRatio_rmse} show that the RMSE associated with non robust methods, i.e. Horn and Gen-Horn increase monotonically. On the contrary, the robust algorithms, GUM-EM and GStudent-EM, have a radically different behavior. After a short increase, the RMSE remains constant, and then it increases again. 

In the remaining experiments, the number of inliers was set to be equal to the number of outliers. 
Figure~\ref{fig:GUaniso_incAmp_rmse} shows the RMSEs for the case when inlier noise is drawn from an anisotropic Gaussian distribution with total variance $\lambda=0.0025$, while outlier noise is drawn from a uniform distribution whose volume is increased from $a=0.2$ to $a=1.0$. Finally, Figure~\ref{fig:GGaniso_incAmp_rmse} shows the RMSEs when inlier noise is drawn from an anisotropic Gaussian distribution with total variance $\lambda=0.0025$, while outlier noise is drawn from an anisotropic Gaussian distribution with total variance varying from $\lambda=0.2$ to $\lambda=1.0$.

\addnote[robust-discussion]{2}{These experiments clearly show that GUM and GStudent can deal with up to 50\% landmarks affected by a substantial noise level (1.5 times the size of the image). The posteriors (GUM) and the weights (GStudent), estimated by Algorithms \ref{algo:em-robfa} and \ref{algo:em-student}, respectively, characterize the observed landmarks and reduce the importance of those landmarks that have large errors in localization. As can be seen from \eqref{eq:posterior-gum}, the GUM posteriors are in the range $0<\alpha<1$ and consequently their values are a relative measure of the importance of the landmarks. In contrast, 
the GStudent weights \eqref{eq:weight-expectation} are in the range $0< w < \infty$, hence they constitute an absolute measure.
Altogether, this offers a valuable framework for building a statistical landmark model from a training dataset.}


\section{3DFA Performance Analysis}
\label{sec:statistical-face}
In this section we describe an unsupervised methodology for quantitatively assessing the performance of 3DFA algorithms. The idea is to apply 3DFA to a dataset of face images in order to extract 3D landmarks, to robustly estimate the rigid transformation that maps these facial landmarks onto a 3D landmark model, and to analyze the discrepancy between the extracted-and-mapped 3D landmarks and the model. Based on a confidence score, it is then possible to decide whether a landmark is correctly localized. This allows to assess the overall performance of a 3DFA algorithm as well as its behavior with respect to various sources of perturbations.

\subsection{Neutral Frontal Landmark Model}
\label{section:neutral-model}
\begin{figure}[t!]
\centering
\subfloat{\includegraphics[width=.11\textwidth,keepaspectratio]{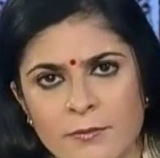}}\hspace{1mm}
\subfloat{\includegraphics[trim=5.5cm 3.5cm 5cm 3.5cm,clip,width=.11\textwidth,keepaspectratio]{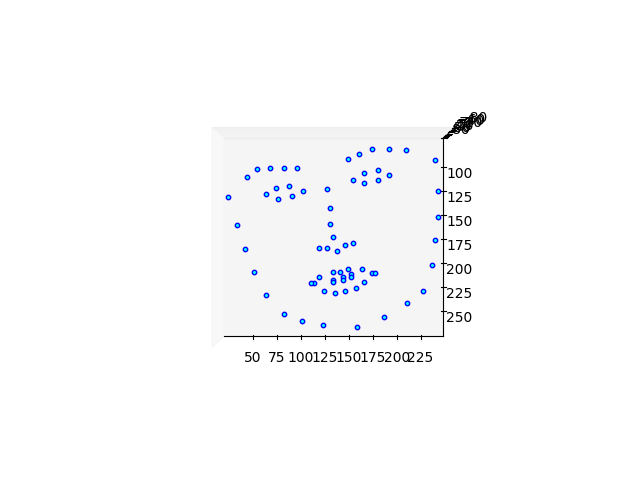}}\hspace{1mm}
\subfloat{\includegraphics[width=.11\textwidth,keepaspectratio]{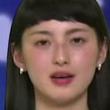}}\hspace{1mm}
\subfloat{\includegraphics[trim=5.5cm 3.5cm 5cm 3.5cm,clip,width=.11\textwidth,keepaspectratio]{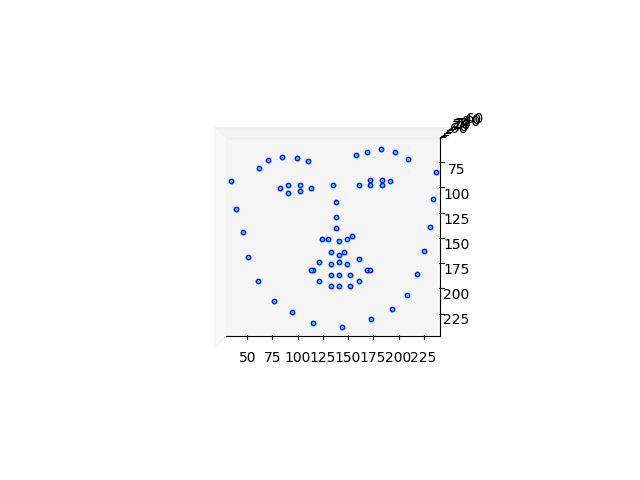}}\hspace{1mm}
 \subfloat{\includegraphics[width=.11\textwidth,keepaspectratio]{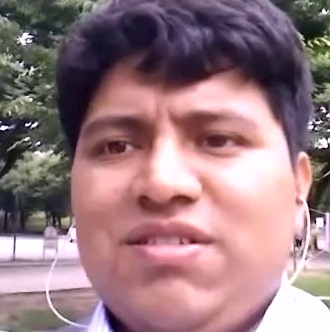}}\hspace{1mm}
 \subfloat{\includegraphics[trim=5.5cm 3.5cm 5cm 3.5cm,clip,width=.11\textwidth,keepaspectratio]{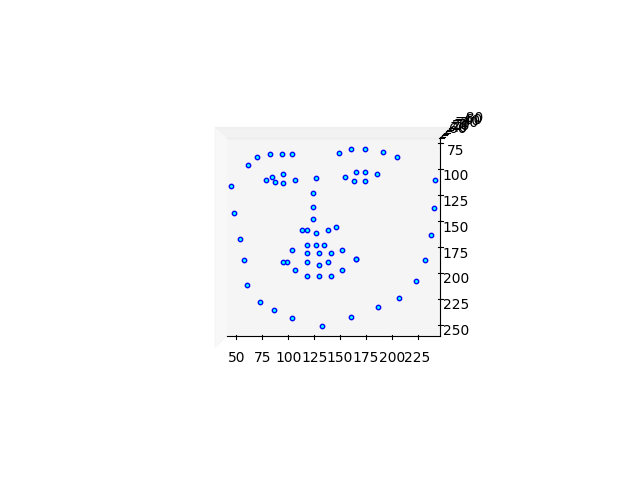}}\hspace{1mm}
\subfloat{\includegraphics[width=.11\textwidth,keepaspectratio]{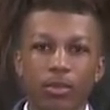}}\hspace{1mm}
\subfloat{\includegraphics[trim=5.5cm 3.5cm 5cm 3.5cm,clip,width=.11\textwidth,keepaspectratio]{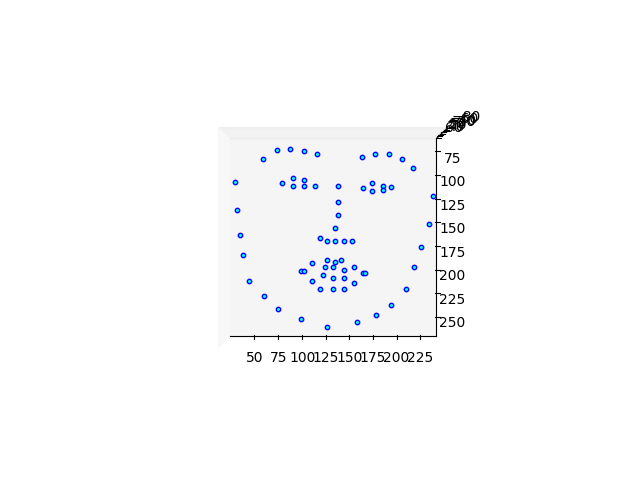}}\hspace{1mm}\\
\subfloat{\includegraphics[trim=2.8cm 1.0cm 3.5cm 1.0cm,clip,width=.35\textwidth,keepaspectratio]{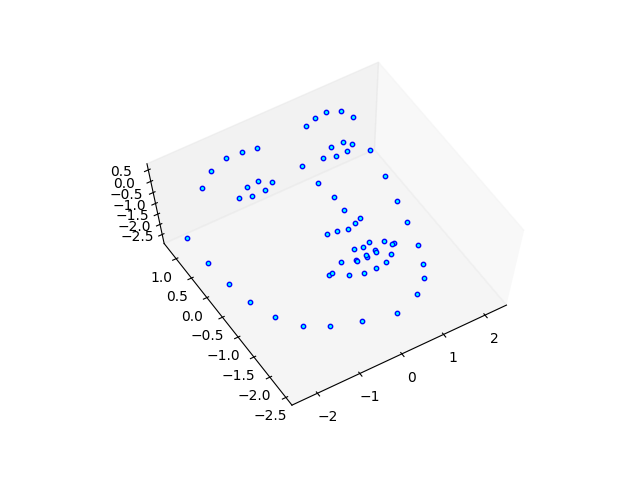}}
\caption{\label{fig:neutral-model-ex} Examples of faces and corresponding 3D landmarks (top) used to compute a neutral frontal landmark model (bottom).}
        \end{figure}

Let's start by computing a \textit{neutral frontal landmark} (NFL) model, $\yvect_{1:N}$, in the following way. \addnote[pca]{2}{A dataset $\mathcal{D}_1$ of $K$ images of neutral faces (frontal viewing, no expression and no interfering object causing occlusion) is collected and $N$ landmarks are extracted from each one of these $K$ faces, $\{\yvect_{1:N,k}\}_{k=1}^{k=K}$. The directions of maximum variance are then computed for each face as to align the faces using these directions. 
Next, mean coordinates for each landmark $n$ are computed, namely:}
\begin{equation}
\label{eq:neutral-means}
\yvect_n = \frac{1}{K} \sum_{k=1}^K \yvect_{n,k}, \forall n, 1\leq n \leq N.
\end{equation}
\addnote[headpose]{3}{
The NFL model was created in-the-wild by harvesting web images and using the face detector of \citep{viola2001rapid} and the head-pose estimator of \citep{drouard2017robust} in order to select frontal faces.} These images were then visually inspected one-by-one to guarantee shape and aspect variabilities as well as neutral facial expressions.This process yields a dataset $\mathcal{D}_1$ composed of $K=1,000$ images. \addnote[dataset1]{2}{The 3DFA method of \citep{bulat2016two}  was used to extract landmarks from each face in the dataset}. Once aligned, a mean for each landmark is computed, using \eqref{eq:neutral-means}. Figure~\ref{fig:neutral-model-ex} shows a few examples of images from the  $\mathcal{D}_1$ dataset, the detected landmarks using \citep{feng2018joint}, and a 3D view of the neutral frontal landmark model.

\subsection{Statistical Frontal Landmark Model}
\label{section:statistical-model}

Now it is explained how a \textit{statistical frontal landmark} (SFL) model is built, namely $\{ \pvect_{1:N}, \Cmat_{1:N}\}$, where $\pvect_{1:N}$ are posterior means and $\Cmat_{1:N}$  are posterior covariance matrices associated with this model. 
The means and covariances are estimated in the following way.   A second dataset $\mathcal{D}_2$ is considered,  namely the YawDD dataset \citep{abtahi2014yawdd}. The faces in this dataset have large variabilities in terms of face shapes, face aspects, head poses and facial expressions, and with no external sources of perturbation such as the presence of interfering objects that may cause occlusions. \addnote[xland]{2}{First, 3D landmarks are extracted from these face images (see below), namely
$ \{ \xvect_{1:N,l}\}_{l=1}^{l=L}$, using either GUM-EM (Algorithm~\ref{algo:em-robfa})  or GStudent-EM (Algorithm~\ref{algo:em-student}) to robustly estimate the rigid transformations between each landmark-set $\xvect_{1:N,l}$ and the the NFL model  $\yvect_{1:N}$. }
Based on this, $L$ mapping parameters are obtained (one for each face $l$): $L$ scale factors, $L$ rotations and $L$ translations:
$\{s_l^{\mathrm{Alg}}, \Rmat_l^{\mathrm{Alg}}, \tvect_l^{\mathrm{Alg}}\}_{l=1}^L$, where the over-script ${\mathrm{Alg}}$ denotes a robust algorithm, namely either GUM-EM or GStudent-EM. \addnote[figure-of-merit]{2}{It is reminded that both algorithms provide a figure of merit characterizing each landmark: 
posterior probabilities $\{\alpha_{n,l}\}_{n=1}^{n=N}$ in the case of GUM-EM, i.e. \eqref{eq:posterior-gum}, and posterior weight means $\{\overline{w}_{n,l}\}_{n=1}^{n=N}$ in the case of GSudent-EM, i.e. \eqref{eq:weight-expectation}. 
Applying one of these robust rigid-alignment methods provides 
frontal landmarks, $\{ \tvv{x}_{1:N,l}^{\mathrm{Alg}}\}_{l=1}^{l=L}$, namely:
\begin{equation}
\tvv{x}_{n,l}^{\mathrm{Alg}} = s_l^{\mathrm{Alg}} \Rmat_l ^{\mathrm{Alg}} \xvect_{n,l} + \tvect_l^{\mathrm{Alg}}. 
\end{equation}
One can now build first- and second-order statistics with the use of posterior probabilities (GUM-EM) and posterior weights (GStudent-EM), respectively:
\begin{align}
\label{eq:GUM-mean}
\pvect_n^{\textrm{GUM}} = \sum_{l=1}^L \alpha_{n,l} \tvv{x}_{n,l}^{\textrm{GUM}} / \sum_{l=1}^L \alpha_{n,l}, 
\end{align}
\begin{align}
\label{eq:GUM-covariance}
\Cmat_n^{\textrm{GUM}} = \frac{\sum_{l=1}^L \alpha_{n,l} ( \tvv{x}_{n,l}^{\textrm{GUM}} - \pvect_n^{\textrm{GUM}} ) ( \tvv{x}_{n,l}^{\textrm{GUM}} - \pvect_n^{\textrm{GUM}} )\tp}{\sum_{l=1}^L \alpha_{n,l}},
\end{align}
and
\begin{align}
\label{eq:GSt-mean}
\pvect_n^{\textrm{GSt}} = \sum_{l=1}^L \overline{w}_{n,l} \tvv{x}_{n,l}^{\textrm{GSt}} / \sum_{l=1}^L \overline{w}_{n,l},  
\end{align}
\begin{align}
\label{eq:GSt-covariance}
\Cmat_n^{\textrm{GSt}} = \frac{ \sum_{l=1}^L \overline{w}_{n,l} ( \tvv{x}_{n,l}^{\textrm{GSt}} - \pvect_n^{\textrm{GSt}} ) ( \tvv{x}_{n,l}^{\textrm{GSt}} - \pvect_n^{\textrm{GSt}} )\tp}{\sum_{l=1}^L \overline{w}_{n,l} }.
\end{align}
}
\begin{figure}[t!]
	\centering
	\subfloat[3DFA1/GUM-EM]{\includegraphics[width=.24\textwidth,keepaspectratio]{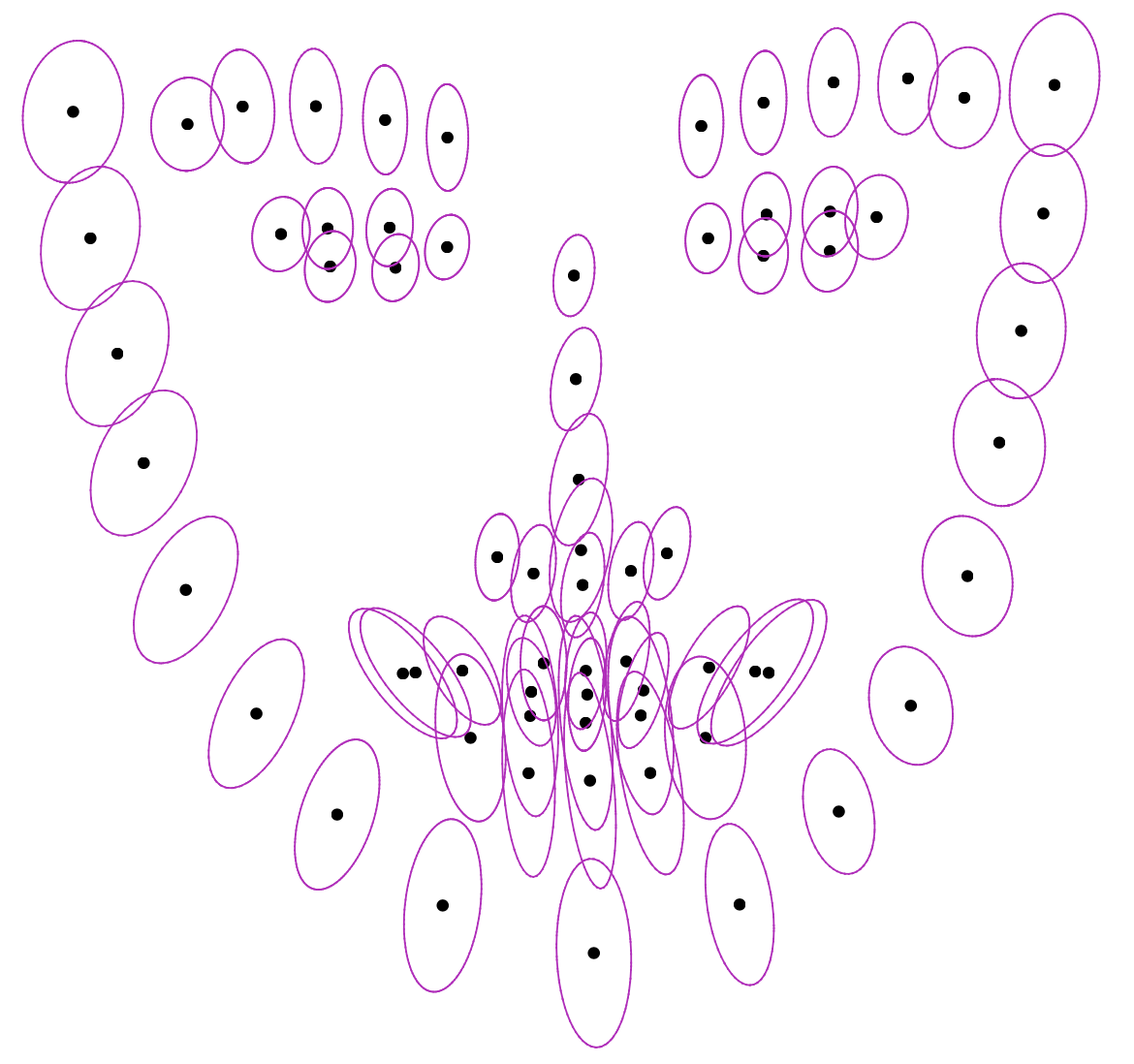}}
	\subfloat[3DFA2/GUM-EM]{\includegraphics[width=.24\textwidth,keepaspectratio]{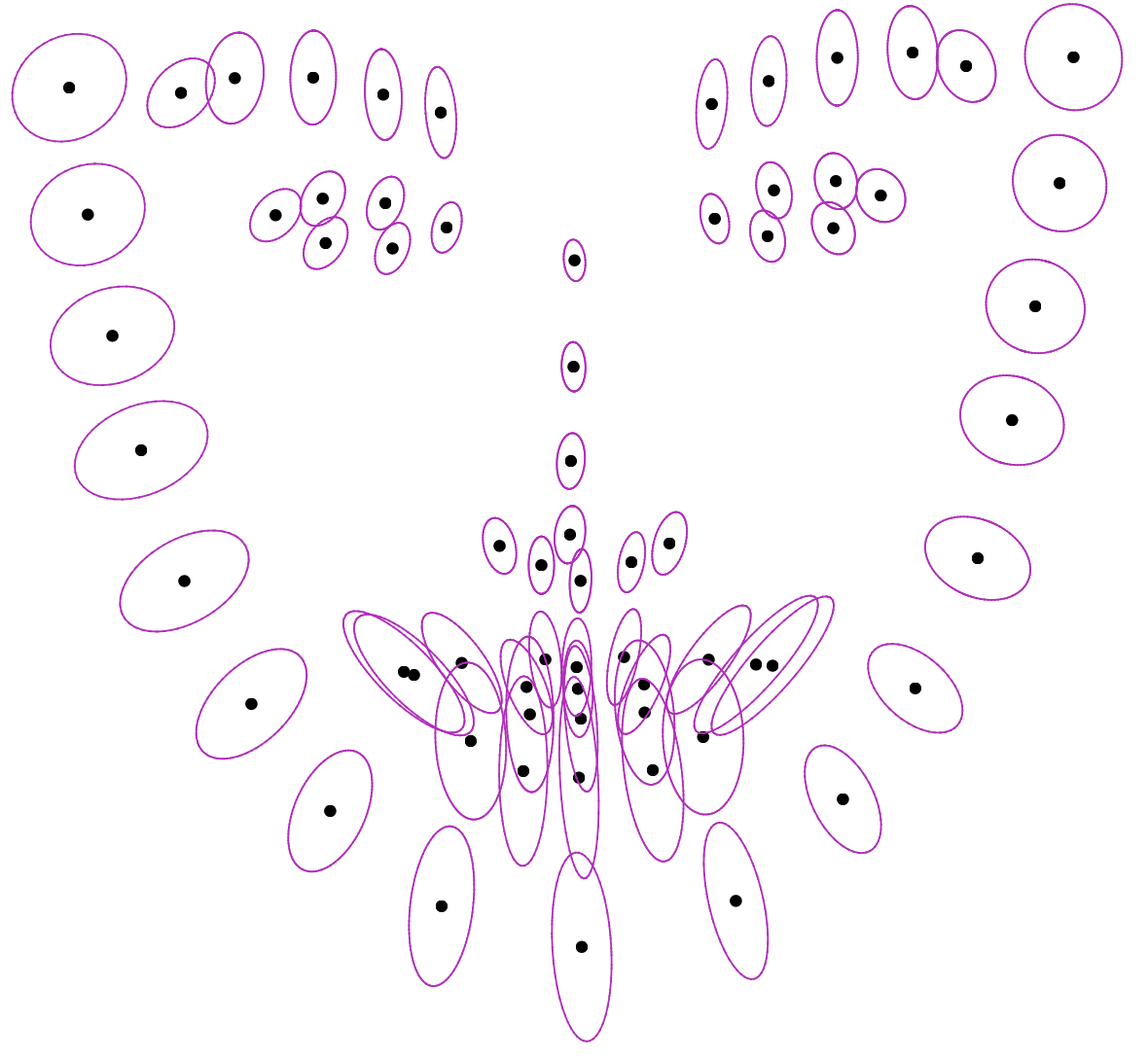}}\\
	\subfloat[3DFA1/GStudent-EM]{\includegraphics[width=.24\textwidth,keepaspectratio]{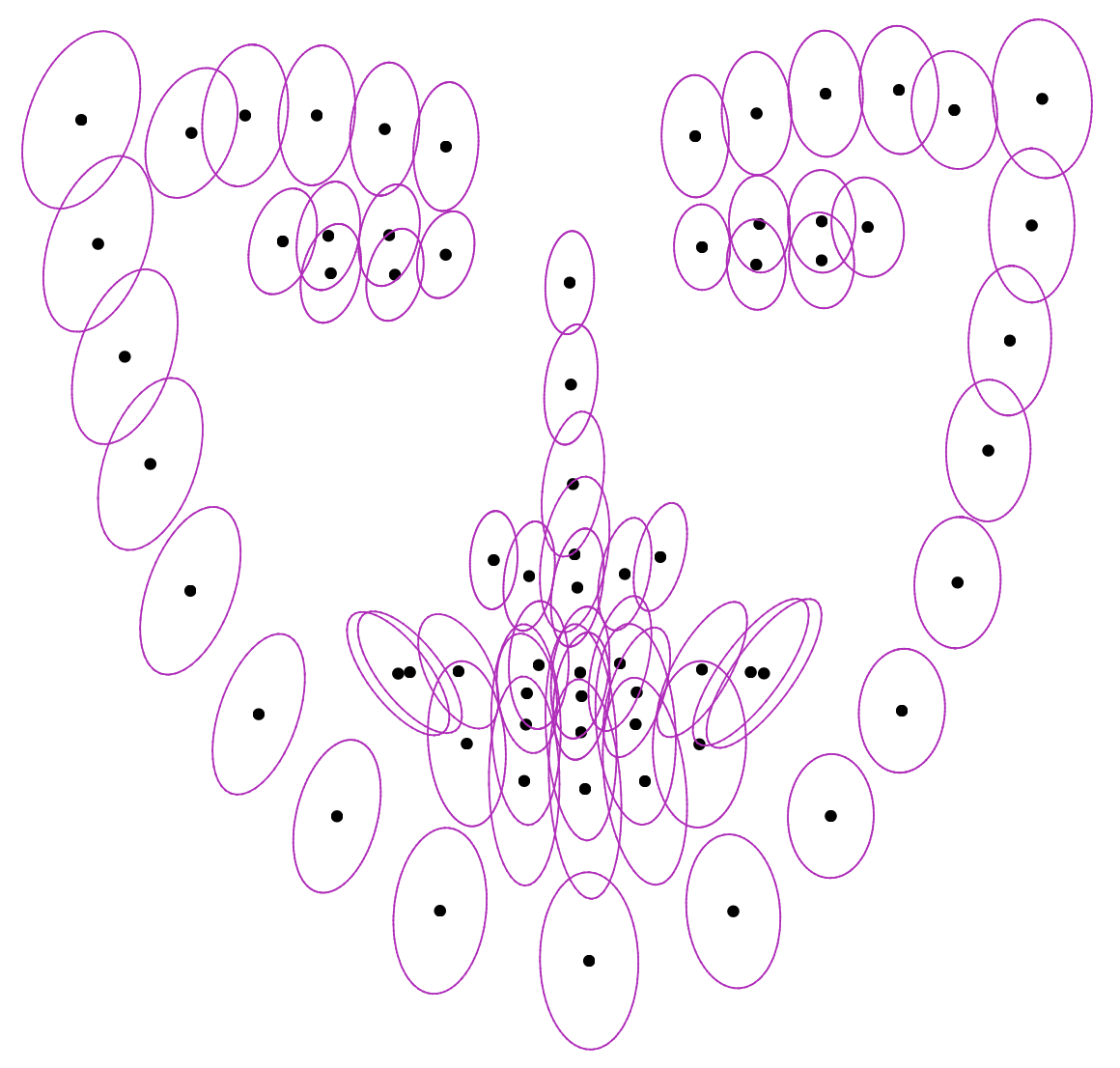}}
	\subfloat[3DFA2/GStudent-EM]{\includegraphics[width=.24\textwidth,keepaspectratio]{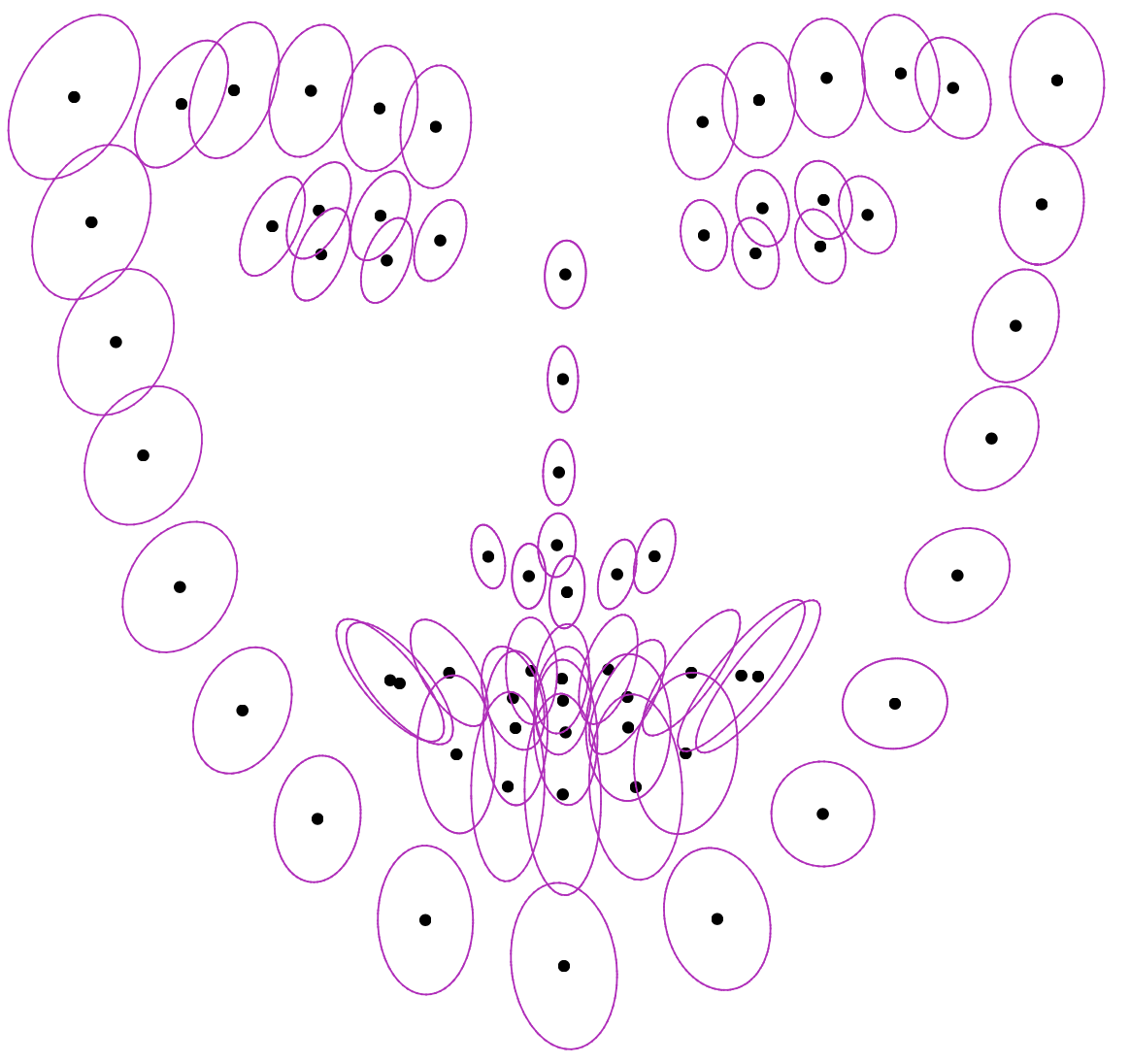}}
	\caption{\label{fig:statistical-model} The four SFL models obtained with two 3DFA methods and with the proposed robust algorithms. The figure shows the image projections of these 3D models.}
\end{figure}

\begin{figure}[t!h!]
\centering
\includegraphics[width=.23\textwidth,keepaspectratio]{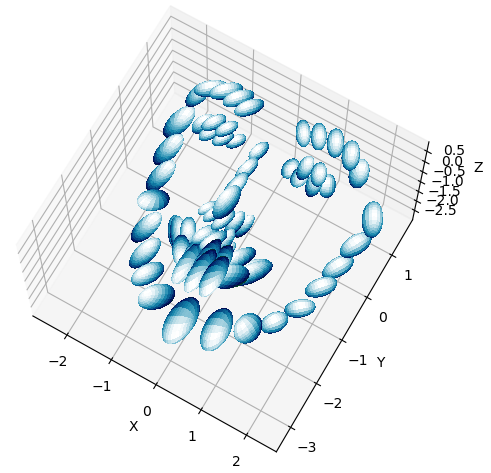}
\includegraphics[width=.23\textwidth,keepaspectratio]{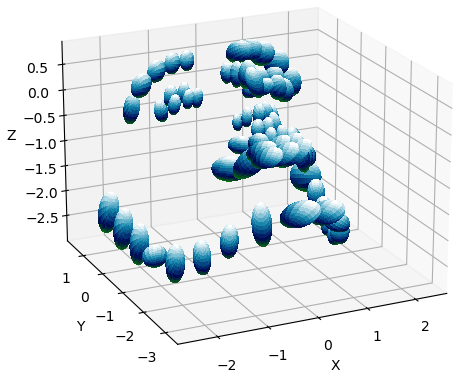}
\caption{\label{fig:statistical-model-3d} Two 3D views of the 3DFA1/GStudent-EM SFL model, i.e. displayed in Fig.~\ref{fig:statistical-model}(c).}
\end{figure}

\addnote[yawdd]{3}{The YawDD dataset \citep{abtahi2014yawdd} contains 342 videos of 107 participants. The videos were recorded at 30~FPS and each video lasts between 15 and 40 seconds, which is equivalent to $L=300,000$ images. }
All the images were processed with no human intervention, namely: face detection \citep{viola2001rapid}, 3D face alignment, and robust rigid alignment with the NFL model just described. This yields the SFL model described above. For that purpose two 3DFA methods were used and the two robust alignment algorithms described in this paper. Hence, there are four possible 3DFA and robust alignment combinations used to train four different models:
\begin{enumerate}
\item 3DFA1/GUM-EM: \citep{bulat2016two} and Algorithm~\ref{algo:em-robfa},
\item 3DFA2/GUM-EM: \citep{feng2018joint} and Algorithm~\ref{algo:em-robfa}, 
\item 3DFA1/GStudent-EM: \citep{bulat2016two} and Algorithm~\ref{algo:em-student}, and
\item 3DFA2/GStudent-EM: \citep{feng2018joint} and Algorithm~\ref{algo:em-student}.
\end{enumerate}
Figure~\ref{fig:statistical-model} shows the SFL models obtained with these four combinations. In this figure, the dots correspond to the posterior means, i.e. \eqref{eq:GUM-mean} and \eqref{eq:GSt-mean}, while the ellipses correspond to image projections of the ellipsoids defined by \eqref{eq:confidence-covariance}. Figure~\ref{fig:statistical-model-3d} shows the 3D ellipsoids associated with the 3DFA1/GStudent-EM model.
Each one of these models may well be viewed as a shape atlas.

\subsection{Unsupervised Confidence Test}
\label{section:conf-test}
Let us now develop an unsupervised (statistical) confidence test for assessing whether the accuracy of a landmark, i.e. its 3D coordinates, is within (inlier) or outside (outlier) an expected range \citep{savage1972foundations,huber2018logical}.
Let us drop the algorithm over-script and let $\Cmat_n = \Qmat_n\Lambdamat_n\Qmat_n\tp$ be the eigen factorization of $\Cmat_n$, where $\Qmat_n$ is an orthonormal matrix and $\Lambdamat_n$ is a diagonal matrix containing the eigenvalues. One can now project each landmark $(n,l)$ onto the space spanned by the three eigenvectors of this matrix:
\begin{equation}
\label{eq:eigen-projection}
\tvv{z}_{n,l} = \Qmat_n\tp (\tvv{x}_{n,l} - \pvect_n).
\end{equation}
Landmark $(n,l)$ is an inlier with $99.7\%$ confidence if $\tvv{z}_{n,l}$ lies inside the ellipsoid whose half-axes are three times the standard deviations, or $3\sqrt{\lambda_n^1}$, $3\sqrt{\lambda_n^2}$, $3\sqrt{\lambda_n^3}$, where $\{\lambda_n^1, \lambda_n^2, \lambda_n^3\}$ are the eigenvalues of $\Cmat_n$, or
\begin{equation}
\label{eq:confidence-interval}
{\tvv{z}_{n,l}}{\tp} \widetilde{\Lambdamat}_n\inverse {\tvv{z}_{n,l}} \leq 1,
\end{equation}
where $\widetilde{\Lambdamat}_n = 9\Lambdamat_n$. Combining \eqref{eq:eigen-projection} and \eqref{eq:confidence-interval}, yields $(\tvv{x}_{n,l} - \pvect_n)\tp \Qmat_n\widetilde{\Lambdamat}_n\inverse\Qmat_n\tp(\tvv{x}_{n,l} - \pvect_n) \leq 1$. With the notation 
\begin{equation}
\label{eq:confidence-covariance}
\widetilde{\Cmat}_n = \Qmat_n\widetilde{\Lambdamat}_n\Qmat_n\tp.
\end{equation}
The $99.7\%$ confidence test writes:
\begin{equation}
\label{eq:99confidence}
\begin{cases}
\textrm{if} \quad \| \tvv{x}_{n,m} - \pvect_n \|_{\widetilde{\Cmat}_n} \leq 1 & (n,m) = \textrm{inlier} \\
\textrm{otherwise} & (n,m) = \textrm{outlier},
\end{cases}
\end{equation}
Based on this confidence test, one can now build an unsupervised  \textit{confidence score} (the higher the better) associated with a sample face $m$, namely:
\begin{equation}
\label{eq:u}
\mathrm{u}(m) = \frac{1}{N} \sum_{n=1}^N \mathcal{I}( \| \tvv{x}_{n,l} - \pvect_n \|_{\widetilde{\Cmat}_n} \leq 1),
\end{equation}
where $\mathcal{I}(\cdot)$ denotes the indicator function. 
\addnote[pose-test]{2}{Notice that \eqref{eq:u} corresponds to the percentage of inliers, i.e. landmarks that, once scaled, rotated and translated, lie inside the confidence volume. Therefore, \eqref{eq:u} can be used to assess whether the pose has been correctly estimated, namely $u\leq 0.5$, or not. Indeed, Section~\ref{sec:benchmark-robust} empirically shows that both the GStudent-EM and GUM-EM yield accurate rigid parameters in the presence of up to 50\% outliers.
One may use the value of the Mahalanobis distance to quantify the degree of confidence, or to increase (or decrease) the size of the ellipsoid in order to provide looser (or stricter) inlier/outlier decisions.} For a test dataset $\mathcal{D}_3$ composed of $M$ samples, one can then compute the unsupervised \textit{mean confidence score} (MCS) over the entire dataset:
\begin{equation}
\label{eq:mu}
\mathrm{U} = \frac{1}{M} \sum_{m=1}^M 
\frac{1}{N} \sum_{n=1}^N \mathcal{I}( \| \tvv{x}_{n,l} - \pvect_n \|_{\widetilde{\Cmat}_n} \leq 1).
\end{equation}
\addnote[mcs]{4}{
It should be noted that this unsupervised confidence test is based on the covariance \eqref{eq:confidence-covariance} associated with the SFL model. The covariance thus computed characterizes small-amplitude noise (inliers) generated by several processes, such as shape variabilities, non-rigid deformations and landmark localization noise. These are indistinguishable by our model. Nevertheless, the proposed performance-analysis method is able to detect large localization errors with 99.7\% confidence. Therefore, it is useful to detect outliers associated with 3D face alignment architectures.
}

\subsection{Supervised Metrics}\label{sec:supervised}

\addnote[supervised]{2}{
Whenever a face dataset comes with annotations, one can use the \textit{normalized mean error} (NME) between the detected landmarks and the annotations, as a standard supervised performance measure \citep{bulat2017far}. The NME associated with face $m$ is defined as follows:
\begin{equation}
	\label{eq:nme}
	 \text{NME}(m)= \frac{1}{N} \sum_{n=1}^{N}  \| \vect{x}_{n,m} - \hat{\vect{x}}_{n,m} \| / d_m,
\end{equation}
where $d_m$ is a normalization factor, which could be the 2D or 3D distance between the eye centers, the size of the face- or of the landmark bounding-box.  
For a set of $M$ faces, one defines  the cumulated error distribution (CED) as a function of  $\varepsilon$ (a user-defined parameter):
\begin{equation}
	\text{CED}(\varepsilon) =\frac{1}{M} \sum_{m=1}^M   \mathcal{I} \Big(\text{NME}(m) \leq \varepsilon \Big).
\end{equation}
For supervised performance analysis, it is common to use the area under curve (AUC) of CED up to a value of $\varepsilon$.
Analogous to the unsupervised metric \eqref{eq:u}, one now defines supervised metrics that count the proportion of inliers (higher the better), for a face $m$ and for $M$ faces:
\begin{align}
\label{eq:gte-new}
\mathrm{s}(m) &= \frac{1}{N} \sum_{n=1}^{N} \mathcal{I} \big( \| \vect{x}_{n,m} - \hat{\vect{x}}_{n,m} \|/d_m \leq \varepsilon \big),\\
\label{eq:mgte}
\mathrm{S} &=\frac{1}{M} \sum_{m=1}^M s(m).
\end{align}
}
\subsection{Unsupervised-Supervised Correlation}

Another interesting metric is the correlation between the unsupervised and supervised metrics. The proposed pipeline is used to reject  erroneous annotations:
\begin{equation}
\label{eq:good-samples}
\mathcal{M}_{\tau}=\{m~|~\hat{u}(m)\ge \tau\}, \quad \mathcal{M}_{\tau} \subset \mathcal{D}_3,
\end{equation}
where $\hat{u}(m)$ is the score \eqref{eq:u} for the annotations of face $m$. The percentage of annotation inliers present in $\mathcal{M}_{\tau}$ increases with $\tau$, at the price of drastically decreasing the number of inlier samples, which in turn lowers down the statistical significance of the resulting scores. The correlation is:
\begin{equation}
\label{eq:corr}
\mathrm{Cor}(\tau) = \frac{ \sum\limits_{m\in\mathcal{M}_{\tau}} (\mathrm{u}(m) - \mathrm{U})(\mathrm{s}(m) - \mathrm{S})}
{ \big(
	\sum\limits_{m\in\mathcal{M}_{\tau}} (\mathrm{u}(m) - \mathrm{U})^2 \sum\limits_{m\in\mathcal{M}_{\tau}} (\mathrm{s}(m) - \mathrm{S})^2
	\big)^{\frac{1}{2}}}
\end{equation}


\section{Experimental Results}
\label{sec:experiments}
\begin{table}[b!]
\caption{\label{table:allmethods} 3D face alignment methods that are analysed, the corresponding citations, and the website of the associated software packages that are publicly available.}
\centering
\begin{tabulary}{0.99\columnwidth}{|C|L|C|}
		\hline
Method & References & Code/year \\
\hline
3DFA1 &  \citep{bulat2017far} (ICCV'17) & \citep{3DFA1}/2020 \\
3DFA2 & \citep{feng2018joint}  (ECCV'18) & \citep{3DFA2}/2018\\
3DFA3 & \citep{zhu2016face} (CVPR'18), \citep{zhu2019face} (PAMI'19) & \citep{3DFA3}/2019 \\
3DFA4  & \citep{tu2dasl19} (IEEE TMM'21) & \citep{3DFA4}/2019 \\
3DFA5 & \citep{guo2020towards} (ECCV'20) & \citep{3DFA5}/2021 \\ \hline
\end{tabulary}
\end{table}

Once the NFL and SFL models are computed using datasets $\mathcal{D}_1$ and $\mathcal{D}_2$, respectively, we use a third dataset, $\mathcal{D}_3$, to empirically assess the performance of five 3DFA algorithms based on the supervised, unsupervised and correlation metrics just described. Table~\ref{table:allmethods} lists the methods that are analysed.

\begin{table*}[t!h!]
\caption{\label{table:unsupervised-metrics} Performance analysis based on MCS computed with  \eqref{eq:mu}. The numbers correspond to the proportion of inliers (the higher the better) over the  AFLW2000-3D dataset, that contains 2,000 face images and 68 landmarks per face. The last two rows show the results of applying the proposed unsupervised metric to the 3D annotations obtained via a semi-automatic process described in \citep{zhu2016face}.}
	\centering
	\begin{tabulary}{0.9\linewidth}{|C|C|C|C|C|C|}
		\hline
		 Test using $\mathcal{D}_3$ & \multicolumn{5}{c|}{Neutral/statistical models computed/trained with $\mathcal{D}_1$/$\mathcal{D}_2$ datasets:}  \\ \cline{2-6}
		 3DFA\#/Method: &  \citep{bulat2017far}/GUM & \citep{feng2018joint}/GUM &  \citep{bulat2017far}/GStudent & \citep{feng2018joint}/GStudent & Mean \\ \hline
		 3DFA1/GUM          & 0.89 & 0.65& 0.93 & 0.80 & 0.82\\ \hline
		3DFA2/GUM           & 0.93 & 0.88 & 0.95 &  0.93 & 0.92\\ \hline
		3DFA3/GUM           & \textbf{0.98} & \textbf{0.96} & \textbf{0.98} &  \textbf{0.98} & \textbf{0.98}\\ \hline
		3DFA4/GUM           & 0.11 & 0.06 & 0.16 &  0.12 & 0.11\\ \hline
		3DFA5/GUM & \textbf{0.98} & 0.95 &\textbf{0.98} & 0.97  & 0.97  \\ \hline
		3DFA1/GStudent      & 0.80  & 0.57 &  0.88 &  0.74  & 0.75 \\ \hline
		3DFA2/GStudent       & 0.84 & 0.76 & 0.90 &   0.88   & 0.84 \\ \hline
		3DFA3/GStudent       & 0.93 & 0.85 & 0.96 &   0.94   & 0.92 \\ \hline
		3DFA4/GStudent       & 0.18 & 0.12 & 0.23 &   0.18   & 0.18 \\ \hline
		3DFA5/GStudent & 0.94 & 0.86 & 0.97 &  0.95   & 0.93 \\ \hline
		ANN/GUM         		& 0.73  & 0.54 & 0.82 &  0.71  & 0.70 \\ \hline
		ANN/GStudent        	 &  0.67 & 0.48 & 0.78 &  0.66  &0.65 \\ \hline
	\end{tabulary}
	\end{table*}

For this purpose AFLW2000-3D \citep{zhu2016face} is used as $\mathcal{D}_3$, consisting of $2000$ images with large pose variations: the yaw angles (vertical axis of rotation) are in the interval $[0^\circ,\pm 30^\circ]$ for 1306 faces, in $[\pm 30^\circ,\pm 60^\circ]$ for 462 faces and in $[\pm 60^\circ, \pm 90^\circ]$
for 232 faces. The dataset contains a large variety of identities, expressions and illumination conditions. Moreover, there are many faces with partial occlusions caused by the presence of hair, hands, glasses, etc. Notice that large poses induce self occlusions. 

\addnote[annot-semi]{3}{Each image in AFLW2000-3D is annotated with 68 3D landmarks. The annotation is performed by fitting a 3DMM to 2D facial landmarks  \citep{zhu2016face}. Failures of this automatic annotation process are manually corrected, hence it is a semi-automatic annotation.\footnote{Please consult \url{https://openaccess.thecvf.com/content_cvpr_2016/supplemental/Zhu_Face_Alignment_Across_2016_CVPR_supplemental.pdf}.}  As noted in \citep{bulat2017far}, in spite of manual correction, annotation errors are still present, especially in the case of profile views. Hence, performance evaluation based on supervised metrics is likely to be biased by these annotation errors. In \citep{bulat2017far} it is qualitatively (visually) shown that in these extreme poses their 3DFA method yields more precise landmark localization results than the semi-automatic annotations. The proposed unsupervised metric is one possible way to quantify the results of \citep{bulat2017far}.}
In an attempt to analyse the semi-automatic annotation process itself, the proposed unsupervised metrics was applied to the annotated landmarks (ANN) of \citep{zhu2016face}, yielding two combinations: ANN/GUM and ANN/GSt.

The results that were obtained based on computing MCS, i.e. \eqref{eq:mu} are summarized in Table~\ref{table:unsupervised-metrics}. It is reminded that $\mathcal{D}_1$ and $\mathcal{D}_2$ were used to compute the NFL model and to train the SFL model, respectively, and $\mathcal{D}_3$ to test their performance. The average scores obtained with the annotated landmarks  (the last two rows  and last column of Table~\ref{table:unsupervised-metrics}), are equal to $0.70$ and to $0.65$, respectively, which seems to confirm that the semi-automatic annotations available with AFLW2000-3D contain a substantial amount of errors and that the 3DFA methods \citep{bulat2016two} and \citep{feng2018joint}, used for training, predict landmark locations that are more accurate than the annotated locations themselves. 

\begin{figure*}[p]
	\centering
	\subfloat[2D landmarks (bounding-box)]{{\includegraphics[height=0.18\linewidth]{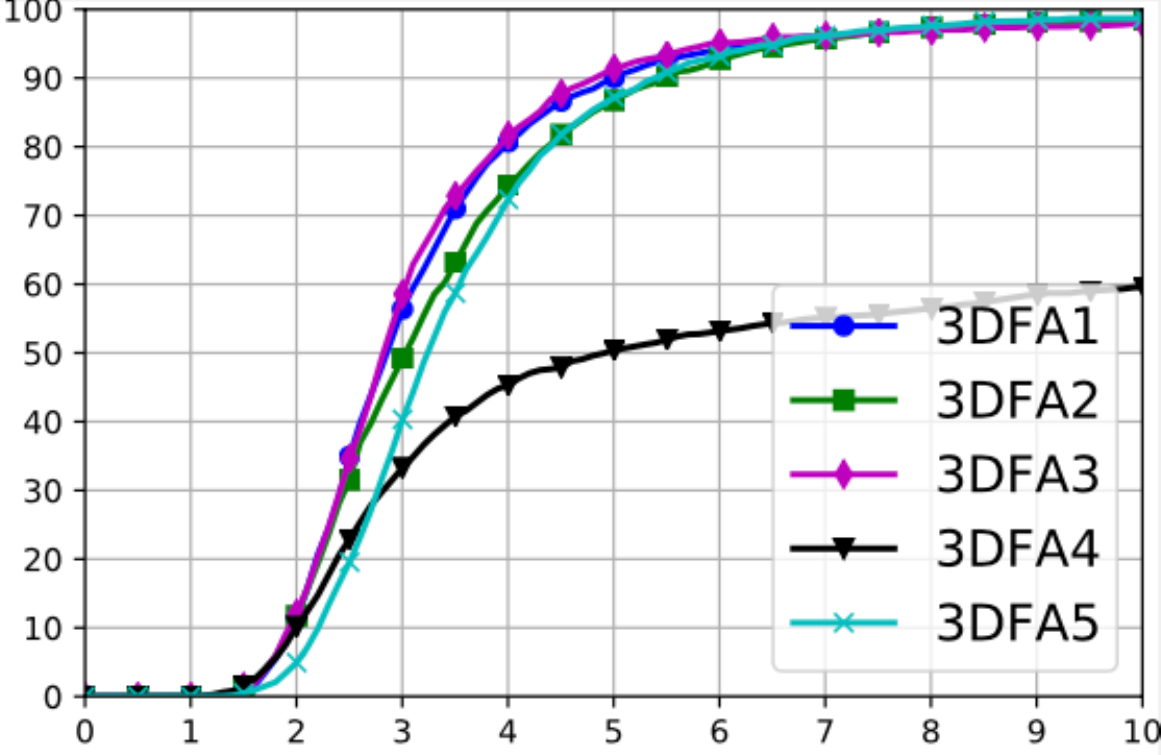} }}
	\subfloat[3D landmarks (bounding-box)]{{\includegraphics[height=0.18\linewidth]{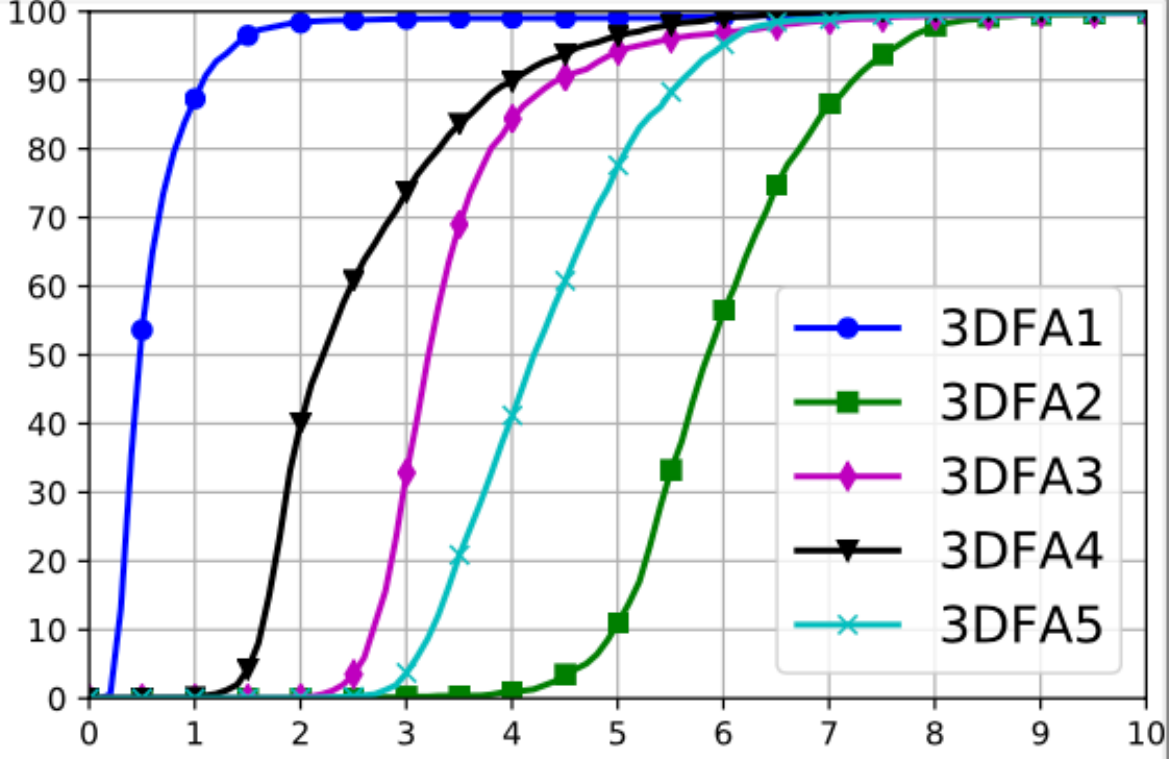} }}
	\subfloat[Centered 3D (bounding-box)]{{\includegraphics[height=0.18\linewidth]{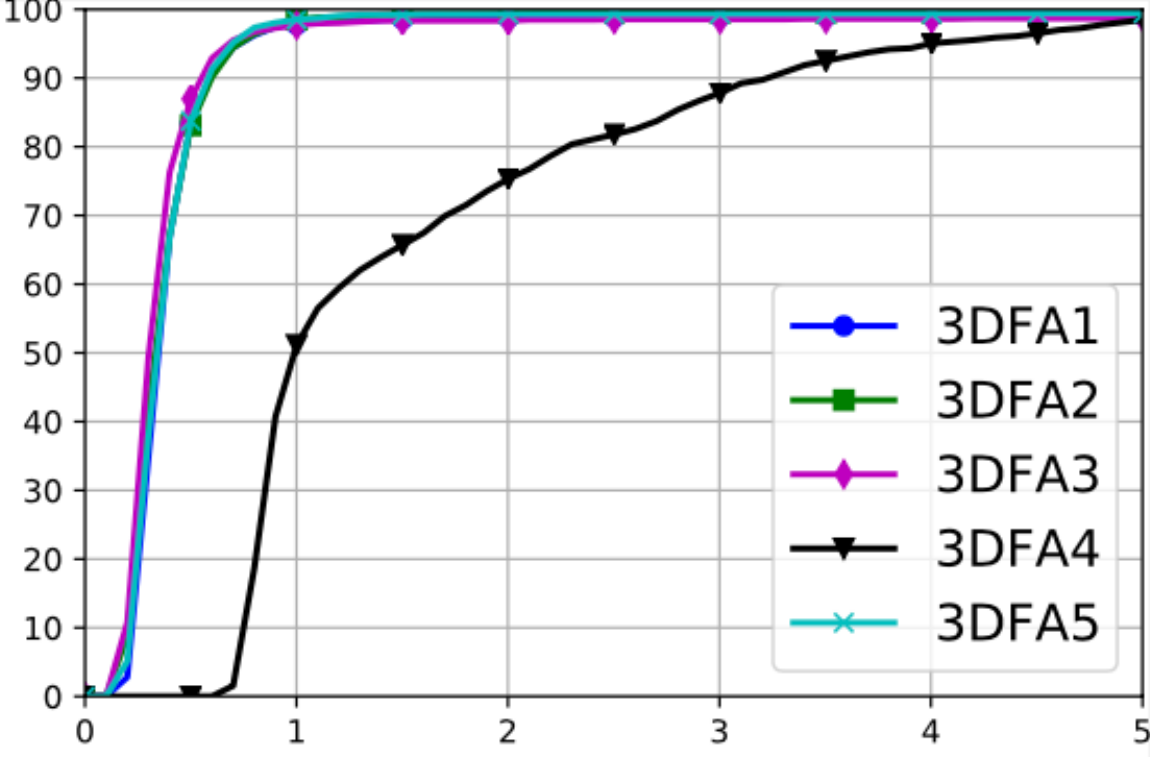} }}\\
	\subfloat[2D landmarks (eye-distance)]{{\includegraphics[height=0.18\linewidth]{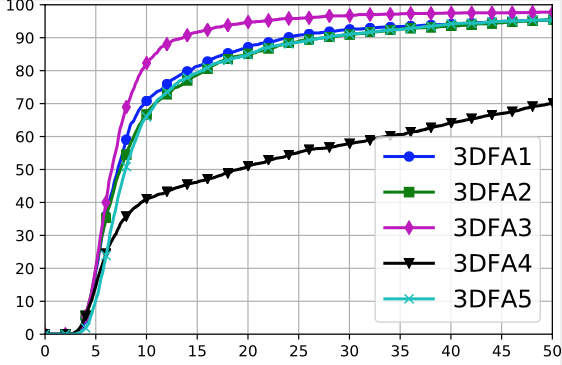} }}
	\subfloat[3D landmarks (eye-distance)]{{\includegraphics[height=0.18\linewidth]{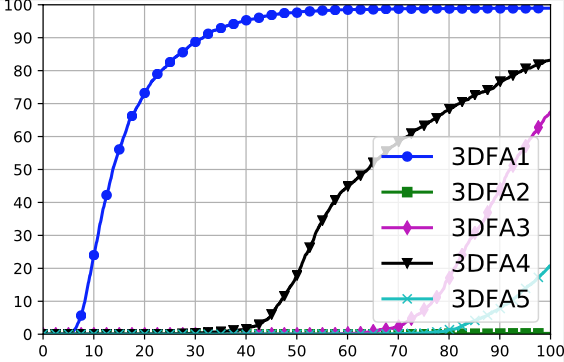} }}
	\subfloat[Centered 3D landmarks (eye-distance)]{{\includegraphics[height=0.18\linewidth]{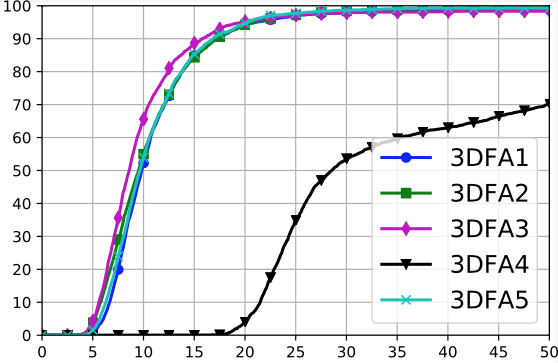} }}
	\caption{\label{fig:new_supervised_metric} Cumulative error distribution (CED) curves computed with \eqref{eq:mgte} for 2D and 3D landmark coordinates, for bounding-box and eye-distance normalizations.}
\end{figure*}

The performance of both 2D and 3D face alignments are analyzed in terms of the supervised metric \eqref{eq:mgte}, where for the 2D case we simply discard depth. In order to remove any bias possibly due to the prediction of  the $z$ coordinate, the performance is also compared by centering this coordinate for all the landmarks of each face. And finally, several NME normalizations are considered (see above). The results are presented in Figure~\ref{fig:new_supervised_metric}. The AUC, NME mean and NME standard deviation, corresponding to Figure~\ref{fig:new_supervised_metric}(d)\&(f), are reported in Table~\ref{table:auc_2dE} and Table \ref{table:auc_3dE}, respectively.

\begin{figure*}[p]
	\centering
	\subfloat[Correlation ($\varepsilon = 0.10$)]{\includegraphics[height=0.18\linewidth]{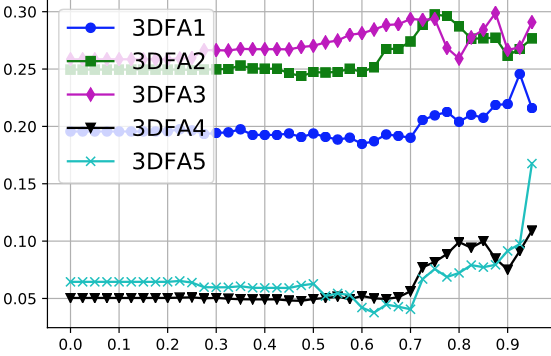} }
	\subfloat[Correlation ($\varepsilon = 0.20$)]{\includegraphics[height=0.18\linewidth]{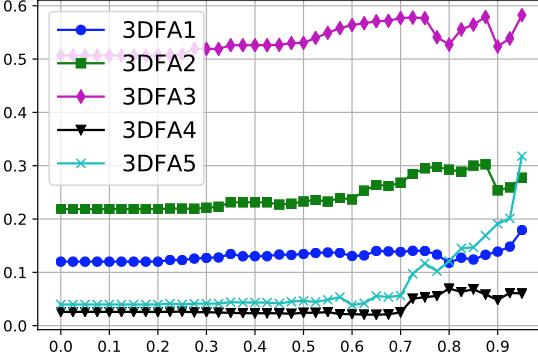} }
	\subfloat[Correlation ($\varepsilon = 0.30$)]{\includegraphics[height=0.18\linewidth]{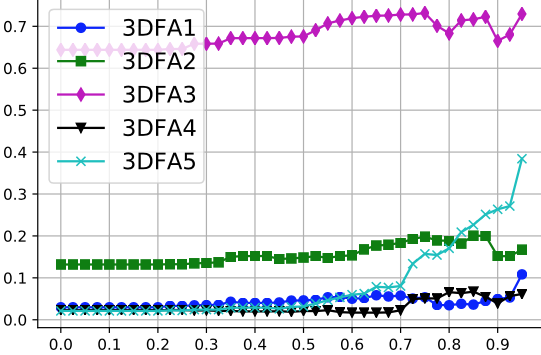} }\\
	\subfloat[P-value ($\varepsilon = 0.10$)]{\includegraphics[height=0.18\linewidth]{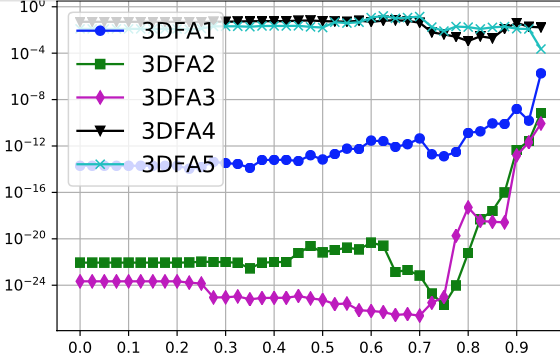} }
   \subfloat[P-value ($\varepsilon = 0.20$)]{\includegraphics[height=0.18\linewidth]{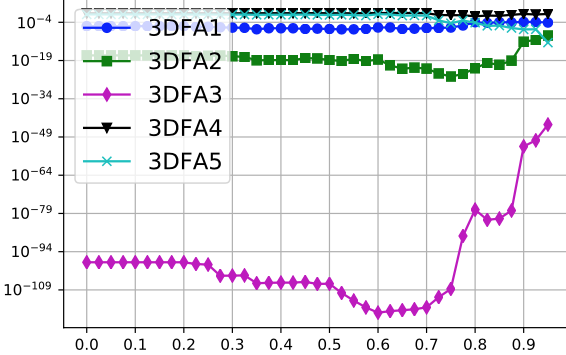} }
  \subfloat[P-value ($\varepsilon = 0.30$)]{\includegraphics[height=0.18\linewidth]{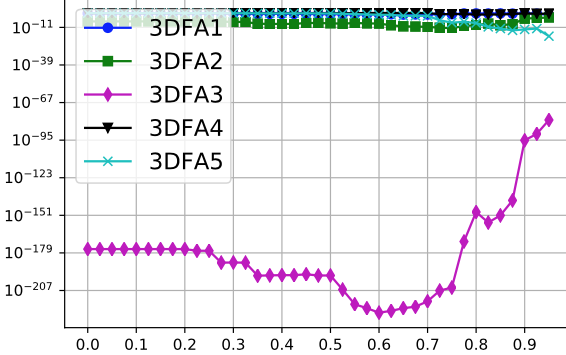} }

	\caption{\label{fig:corr} Correlation (top row) and the corresponding p-value (bottom row) between the supervised and unsupervised metrics for three different values of  $\varepsilon$ defined in \eqref{eq:gte-new}.}
\end{figure*}

In addition, the relation between the supervised metric, i.e. \eqref{eq:gte-new} (with the settings of Figure~\ref{fig:new_supervised_metric}(f)), and the unsupervised metric are analyzed: the correlation coefficients and the corresponding p-values are plotted in Figure~\ref{fig:corr}. The statistical model has been obtained using 3DFA1/GUM. Each figure corresponds to a different value for the threshold $\varepsilon$ in \eqref{eq:gte-new}. As can be seen, increasing the threshold leads to a higher correlation between the supervised and unsupervised metrics.

\begin{figure*}[p]
	\centering
	\subfloat[Results obtained with 3DFA1 \citep{bulat2016two}]
	{
	\includegraphics[width=.42\textwidth]{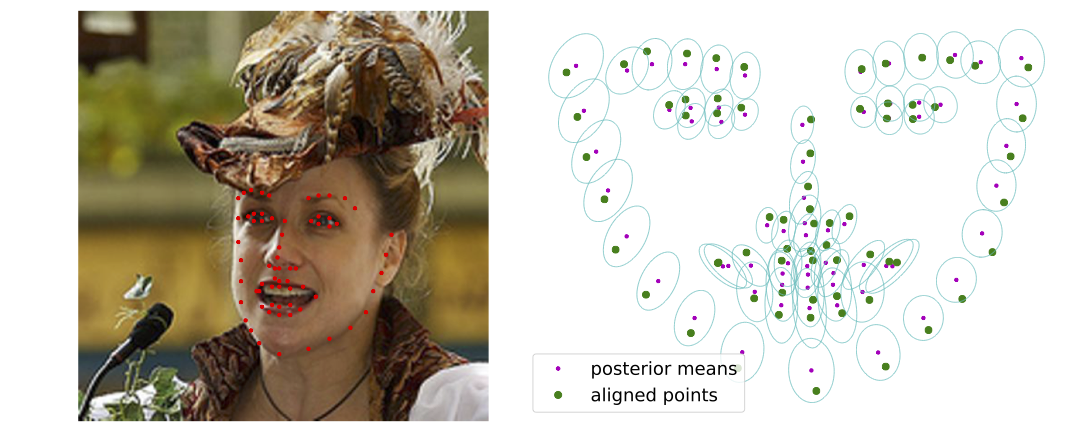} 
	\includegraphics[width=.42\textwidth]{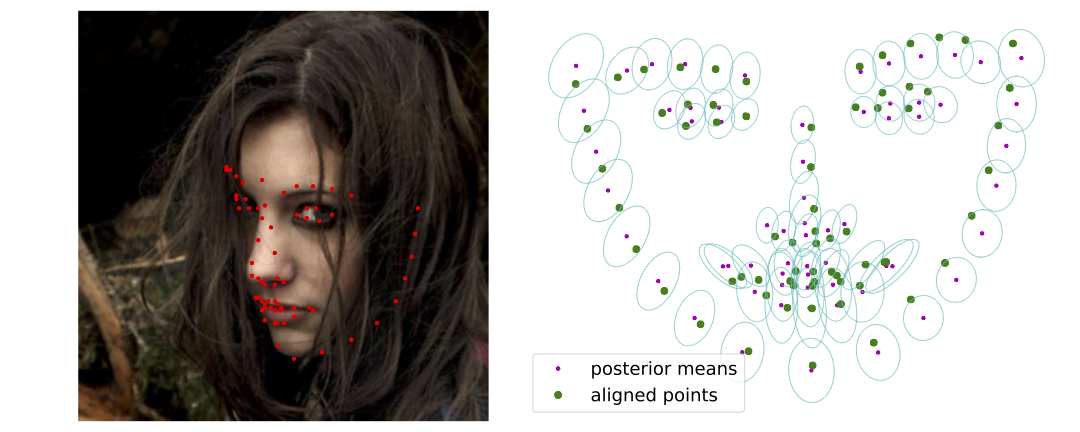}
	} \\
	\subfloat[Results  obtained with 3DFA2 \citep{feng2018joint}]
	{
	\includegraphics[width=.42\textwidth]{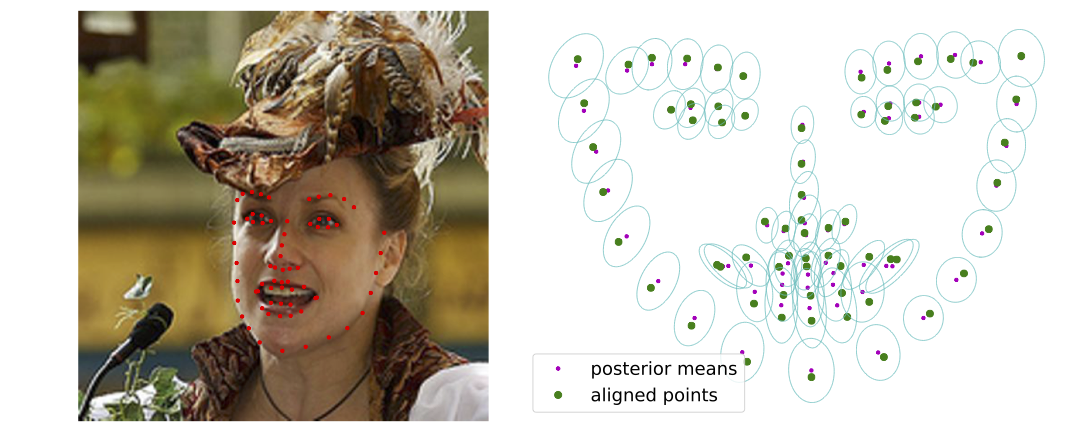}
	\includegraphics[width=.42\textwidth]{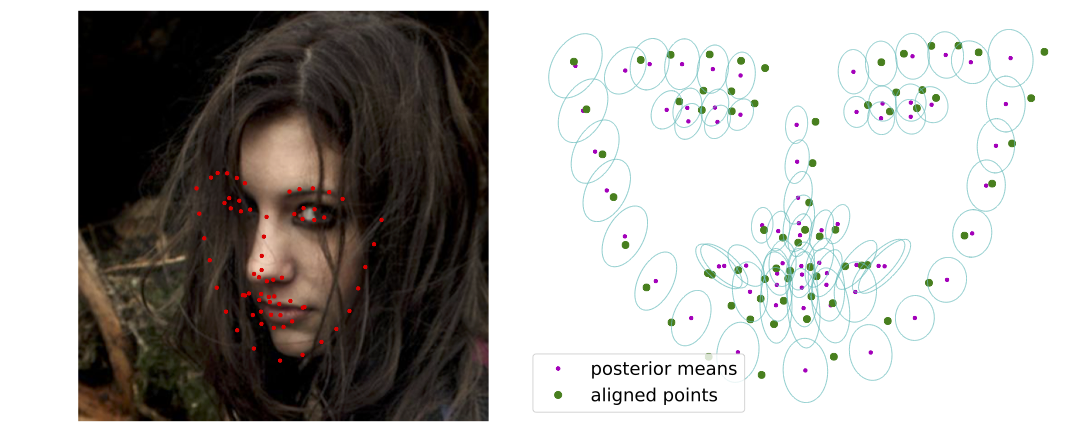}
	} \\
	\subfloat[Results  obtained with 3DFA3 \citep{zhu2016face}]
	{
	\includegraphics[width=.42\textwidth]{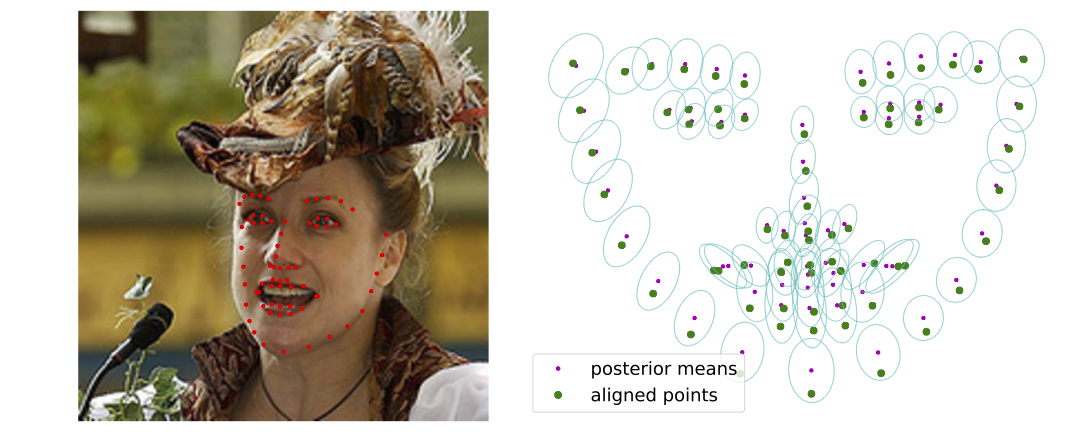} 
	\includegraphics[width=.42\textwidth]{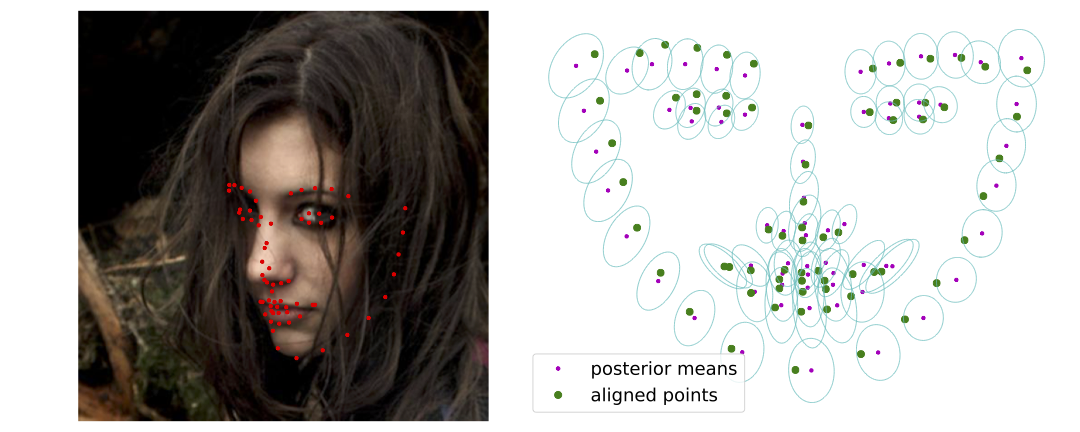}
	} \\
	\subfloat[Results  obtained with 3DFA4 \citep{tu2dasl19}]
	{
	\includegraphics[width=.42\textwidth]{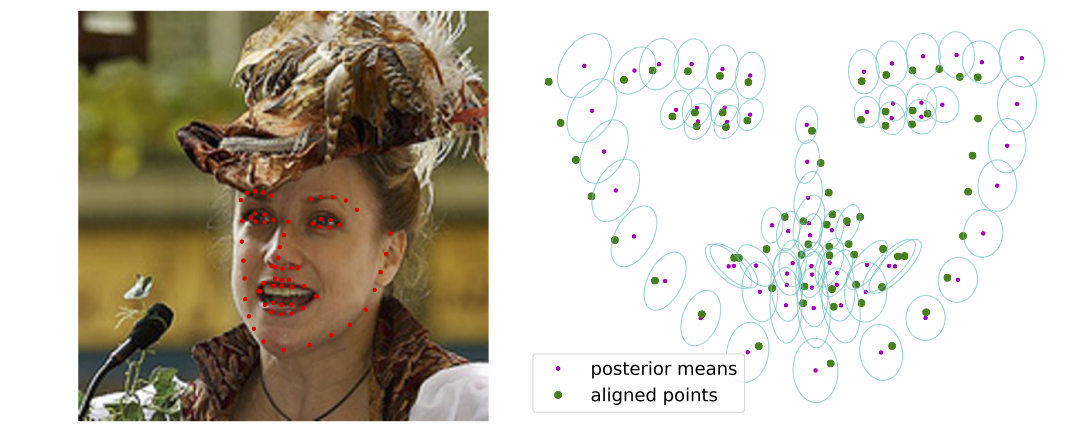}
	\includegraphics[width=.42\textwidth]{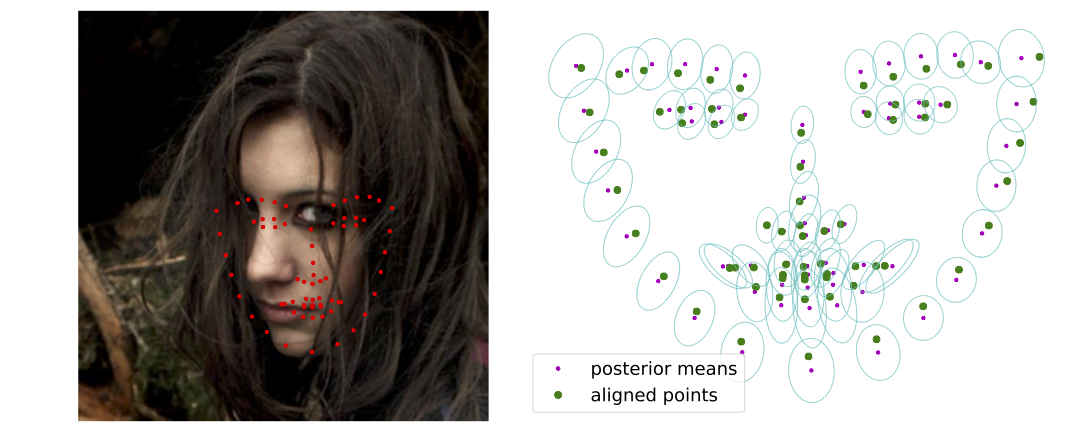}
	} \\
	\subfloat[Results  obtained with 3DFA5 \citep{guo2020towards}]
	{
	\includegraphics[width=.42\textwidth]{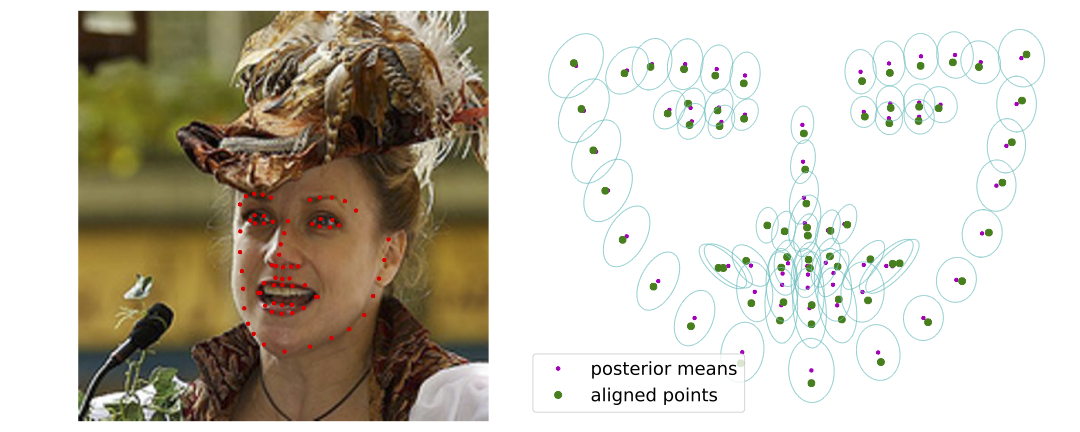}
	\includegraphics[width=.42\textwidth]{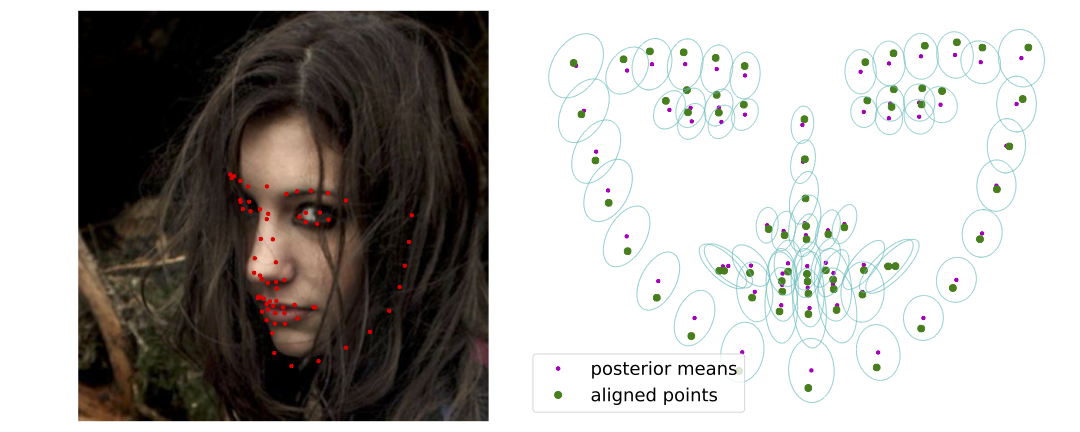}
	} \\
	\subfloat[Results obtained with the semi-automatic annotations  \citep{zhu2016face}.]
	{
	\includegraphics[width=.42\textwidth]{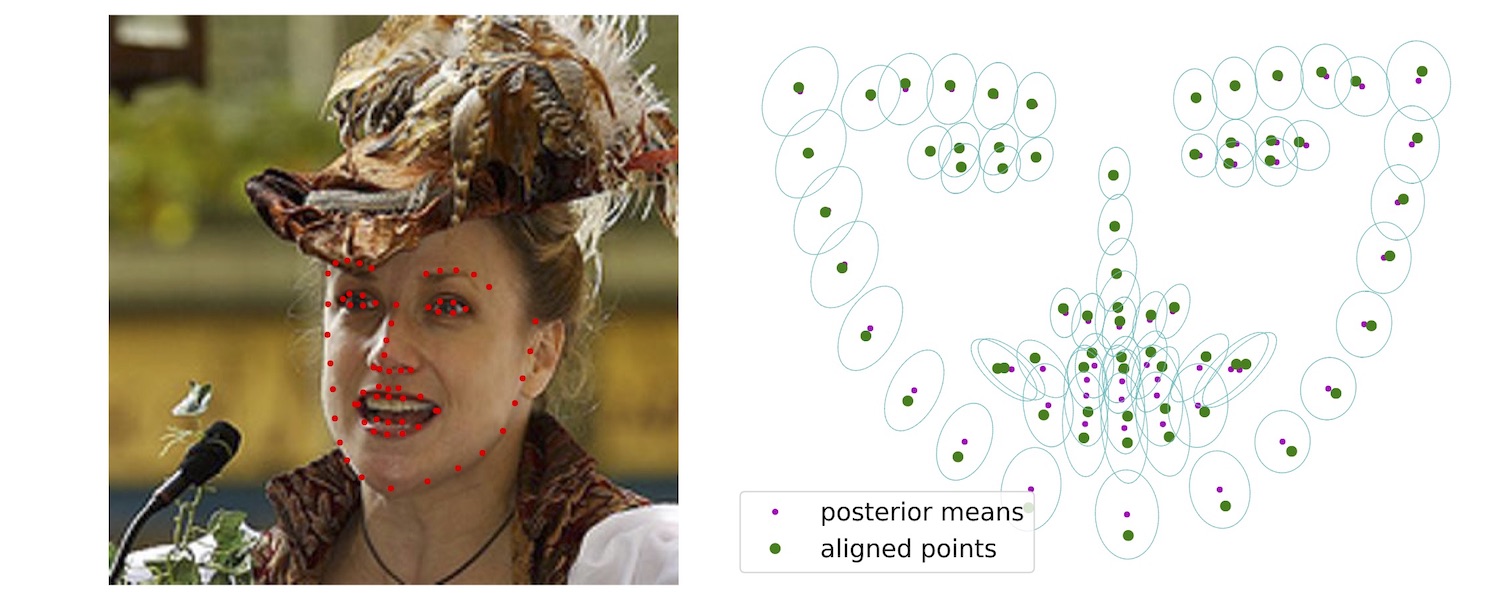} 
	\includegraphics[width=.42\textwidth]{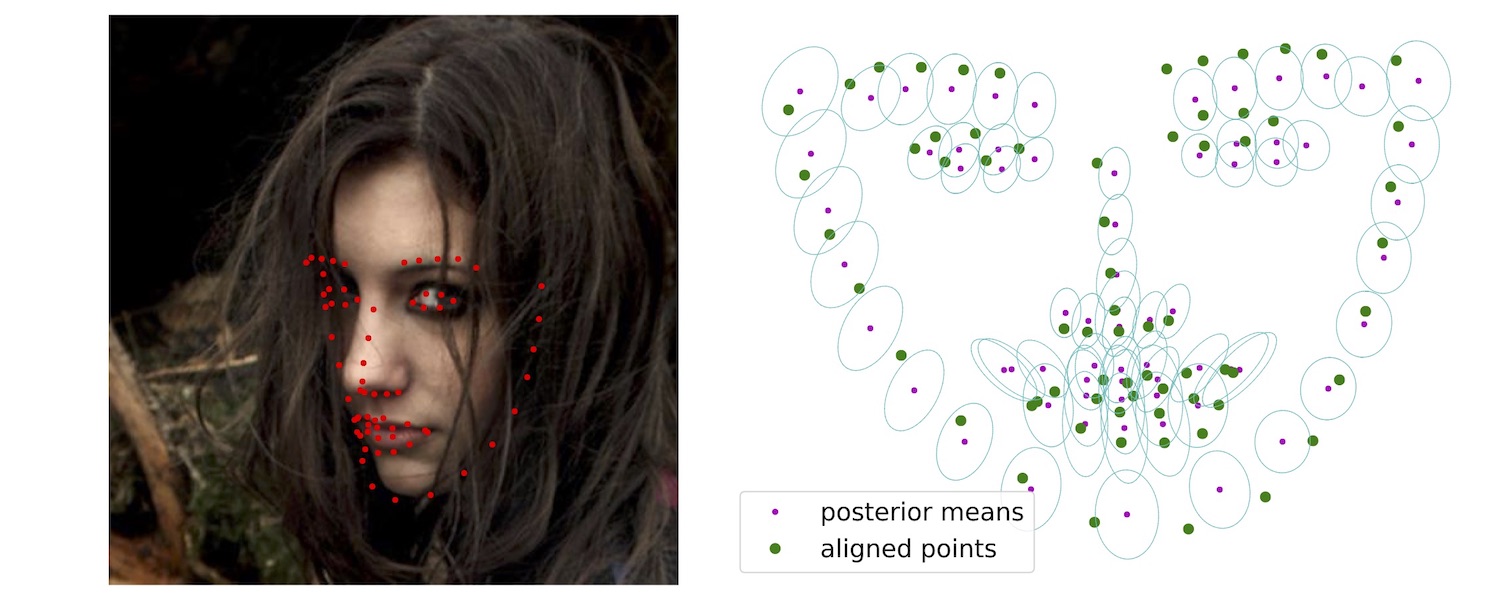} 
	} 
	\caption{\label{fig:3dfa1-examples} Two examples from the AFLW2000-3D dataset \citep{zhu2016face}. Left column: 3D landmarks predicted with five 3DFA architectures and with the semi-automatic annotation method of \citep{zhu2016face} (last row). Right column: results obtained by robustly mapping the predicted landmarks onto the statistical frontal landmark model.}
\end{figure*}

Figure~\ref{fig:3dfa1-examples} illustrates the proposed method with two examples from AFLW2000-3D. In all these examples 
3DFA1/GStudent-EM was used for training and GStudent-EM was used for testing. 
\addnote[limitations]{2}
{One may notice that all the 3DFA methods partially fail on the right hand-side example. Because the predicted results correspond to valid facial landmarks, the proposed pipeline incorrectly classifies many landmarks as inliers.
It is worth noticing that the incorrect landmarks, associated with the automatic annotation process, are correctly classified as outliers. This example shows the limitations of both the unsupervised (proposed) and supervised metrics because they are unable to assess whether the predicted landmarks are coherent with the RGB information.
}

\begin{table}[t!h!]
\caption{\label{table:auc_2dE} AUC, NME mean and NME standard deviation associated with Figure~\ref{fig:new_supervised_metric}(d) up to  NME = 30\%.}
	\centering 
		\begin{tabulary}{0.99\columnwidth}{| L | C | C | C |} 
			\hline
			Method&   AUC $\uparrow$ (\%) & Mean NME $\downarrow$ (\%) & Std. NME $\downarrow$ (\%)\\ 
			\hline
			3DFA1  &  56.50 & 14.87 & 1.67 \\ 
			\hline
			3DFA2  & 57.48  & \textbf{9.22} &  \textbf{0.64} \\ 
			\hline
			3DFA3  &  \textbf{58.24} & 9.84  &   0.95 \\ 
			\hline
			3DFA4  &  35.84 & 31.24 &  1.07 \\ 
			\hline
			3DFA5  &  55.73 & 10.18 &  0.76 \\ 			
			\hline 
	\end{tabulary}
		
\end{table}

\begin{table}[t!h!]
\caption{\label{table:auc_3dE} AUC, NME mean and NME standard deviation associated with Figure~\ref{fig:new_supervised_metric}(f) up to  NME = 30\%.}
	\centering 
		\begin{tabulary}{0.99\columnwidth}{| L | C | C | C |} 
			\hline
			Method&   AUC $\uparrow$ (\%) & Mean NME $\downarrow$ (\%) & Std. NME $\downarrow$ (\%)\\ 
			\hline
			3DFA1  &  49.48 & 11.29 &  0.46 \\ 
			\hline
			3DFA2  & \textbf{52.01}  & \textbf{10.20} &  \textbf{0.34} \\ 
			\hline
			3DFA3  & 51.08 & 12.72 &  0.67\\ 
			\hline
			3DFA4  &  6.00 & 41.67 &  0.71 \\ 
			\hline
			3DFA5  &  51.10 & 10.79 &  0.43 \\ 			
			\hline 
	\end{tabulary}
		
\end{table}

\section{Discussion and Conclusions}
\label{sec:conclusions}

\addnote[conclusions-1]{2}
{
This paper proposes to analyse the performance of DNN-based 3DFA in an unsupervised way. The rationale of the approach is to build a statistical shape model that encodes variabilities due to identities and to non-rigid facial deformations. The performance analysis per-se takes as input a set of predicted 3D landmarks and then maps this set onto a frontal pose.  A statistical confidence test is used to classify each landmark as either an inlier or an outlier. The inlier/outlier decision is based on simply verifying whether a landmark lies inside or outside an ellipsoidal-shaped volume of confidence. Each ellipsoid has 
a posterior mean as a center, and corresponds to a posterior covariance. The use of robust probability distribution functions (Gaussian-uniform mixture or generalized Student t-distribution) enable the proposed method to disregard large errors from the shape model. Indeed, there is a score (or a weight) associated with each landmark, which prevents the confidence volume to grow exaggeratedly large, i.e. \eqref{eq:GUM-covariance} and \eqref{eq:GSt-covariance}.
}

\addnote[conclusions-2]{2}{
The foundational principle of the proposed method is to rotate, translate and scale the landmarks in order to obtain a normalized representation (or a natural mathematical home) for the shape embedded in the facial landmarks. Key to the success of the proposed pipeline is the robust estimation of the rigid parameters: both GUM-EM and GStudent-EM yield accurate rigid parameters in the presence of non-rigid facial deformations and of (up to) 50\% large errors.  This reveals, as expected, that 
landmarks associated with deformable regions of the face, e.g. the lips and the lower jaw, have larger confidence volumes than landmarks associated with rigid regions, e.g. the nose, the eyes and the upper jaw, e.g. Figure~\ref{fig:statistical-model}. The proposed method could also be seen as a procedure to separate rigid head motions from non-rigid facial deformations and, hence, to enable the analysis of facial expressions in realistic settings, i.e. in the presence of head movements. The statistical model could be used as a prior for the dynamic analysis of facial expressions. The method could be easily adapted to the task of evaluating 3D landmarks predicted from 3D face scans, e.g. \citep{zhang2020deep,wang2022learning}.
}

\addnote[limitation]{4}{
A limitation of the proposed methodology is linked to the discriminative nature of 3DFA training and its inherent use of semi-automatic annotations, which is likely to induce errors. The negative effects due to the propagation of these errors are mitigated by the use of computationally tractable robust estimators. Nevertheless, small annotation errors are indistinguishable from variabilities due to identity and expressions. 
}

\addnote[gt-idea]{4}{
The proposed unsupervised analysis empirically reveals that neither 3DFA predictions nor automatic annotations could count as ground-truth landmark coordinates. A promising follow-up is to consider several ``experts" as proposed in \citep{akhondi2014logarithmic}. Here an expert is a 3DFA architecture combined with a robust rigid-mapping estimator. Each such expert could infer a statistical frontal landmark model which may well be  viewed as an observation of the unknown ground truth. Then, one may consider several experts simultaneously and cast the problem of estimating the unknown ground-truth into the problem of maximization of the joint distribution of the complete data, namely the observed data (provided by the experts) and the hidden data (the unknown annotations).
}

\appendix
\section{Closed-Form Solution Using Unit Quaternions}
\label{app:rotation-unitquaternion}
Consider \eqref{eq:expectation-rotation-scale} with $\Sigmamat=\sigma \Imat$. 
We immediately obtain the following formulas for the model parameters:
\begin{align}
\label{eq:optimal-scale-simple}
s^{\star} & = \left(
\frac
{\sum_{n=1}^{N} \alpha_n  \hatvec{y}_n \tp \hatvec{y}_n }
{\sum_{n=1}^{N}  \alpha_n \hatvec{x}_n \tp  \hatvec{x}_n }
\right)^{1/2}. \\
\label{eq:optimization-rotation-quaternion}
\Rmat^{\star} & = \argmin_{\Rmat} \frac{1}{2} \sum_{n=1}^{N} \alpha_n \| \hatvec{y}_n - s^{\star} \Rmat \hatvec{x}_n \|^2,\\
\label{eq:isotropic-cov}
\sigma^{\star} & = \frac{1}{3\sum_{n=1}^{N}\alpha_n} \sum_{n=1}^{N} \alpha_n \| \hatvec{y}_n - s^{\star} \Rmat^{\star} \hatvec{x}_n \|^2,
\end{align}
The formula for the posteriors becomes:
\begin{equation}
\alpha_n = \frac{p(2\pi\sigma)^{-3/2} \exp (-\|\xvect_n\|^2/2\sigma)}{p(2\pi\sigma)^{-3/2} \exp (-\|\xvect_n\|^2/2\sigma) + (1-p)\gamma\inverse}
\end{equation}
It is well known that a rotation matrix can be parameterized by a unit quaternion \citep{horn1987closed}. Let $\Rmat$ be parameterized by its axis and angle of rotation, $\nvect=(n_1  \;  n_2  \;  n_3)\tp$, $\|\nvect\|=1$ and $\phi$. The unit quaternion parameterizing the rotation is:
\begin{align}
q &= \cos \frac{\phi}{2} + \sin \frac{\phi}{2} (i n_1 + j n_2 + k n_3) \nonumber \\
   &= q_0 + i q_1 + j q_2 + k q_3,
\end{align}
with $i^2=j^2=k^2=ijk = -1$, $\qvect = (q_0 \; q_1 \; q_2 \; q_3)\tp\in\mathbb{R}^4$ by abuse of notation, and $\qvect\qvect\tp=1$. 
A vector $\avect\in\mathbb{R}^3$ can be represented as a purely imaginary quaternion, namely $\widetilde{\avect} = (0 \; a_1 \; a_2 \; a_3)\tp\in\mathbb{R}^4$.
The action of a rotation onto $\widetilde{\avect}$ can be written as $\qvect \ast \widetilde{\avect} \ast \ovec{q}$,
where the symbol $\ast$ corresponds to the quaternion product and $\ovec{q}$ is the conjugate of $\qvect$, namely $\overline{q} = q_0 - i q_1 - j q_2 - k q_3$. Making use of the properties $\| \qvect_1 \ast  \qvect_2\|^2 = \| \qvect_1 \|^2 \|  \qvect_2\|^2$ and  $\ovec{q} \ast \qvect = \| \qvect \|^2=1$, the squared Euclidean norm in \eqref{eq:optimization-rotation-quaternion} can be successively written as:
\begin{align}
\| \hatvec{y}_n -  s \Rmat \hatvec{x}_n \|^2 & = \| \tvv{y}_n -  s \qvect \ast  \tvv{x}_n  \ast \ovec{q}\|^2 \| \qvect \|^2 \nonumber \\
&=  \| \tvv{y}_n \ast \qvect -  s \qvect \ast  \tvv{x}_n \ast \ovec{q} \ast \qvect \|^2 \nonumber  \\
&= \| \tvv{y}_n \ast \qvect -  s \qvect \ast  \tvv{x}_n  \|^2 \nonumber \\
& = \| Q(\tvv{y}_n) \qvect  - s W( \tvv{x}_n) \qvect \|^2 \nonumber \\
&= \qvect\tp \Mmat_n \qvect,
\end{align}
with:
\begin{equation}
\label{eq:measurment-M_n}
\Mmat_n =  \big( Q(\tvv{y}_n) - s W( \tvv{x}_n)   \big)\tp \big( Q(\tvv{y}_n) - s W( \tvv{x}_n) \big),
\end{equation}
and where we replaced the quaternion products  $\tvv{a} \ast \qvect$ and $\qvect \ast  \tvv{a}$ with matrix-vector products $\Qmat(\tvv{a}) \qvect $ and $\Wmat(\tvv{a}) \qvect $ , with:
\begin{align}
\Qmat(\tvv{a}) &=
\begin{pmatrix}
0 & -a_1 & -a_2 & -a_3 \\
a_1 & 0 & -a_3 & a_2 \\
a_2 & a_3 & 0 & -a_1 \\
a_3 & -a_2 & a_1 & 0
\end{pmatrix} 
\\
\Wmat(\tvv{a}) &=
\begin{pmatrix}
0 & -a_1 & -a_2 & -a_3 \\
a_1 & 0 & a_3 & -a_2 \\
a_2 & -a_3 & 0 & a_1 \\
a_3 & a_2 & -a_1 & 0
\end{pmatrix} 
\end{align}
Consequently, the right-hand side of \eqref{eq:optimization-rotation-quaternion} writes 
$$\sum_{n=1}^{N} \big( \qvect\tp \alpha_n \Mmat_n \qvect \big) = \qvect\tp \big(\sum_{n=1}^{N} \alpha_n \Mmat_n \big) \qvect =\qvect\tp \Mmat \qvect, $$
where $\alpha_n \geq 0$ and $\Mmat_n\in\mathbb{R}^{4\times 4}$ is semi-definite positive symmetric, i.e. \eqref{eq:measurment-M_n}, hence so is $\Mmat$.
By constraining the minimizer to be a unit quaternion, we obtain the following minimization problem:
\begin{equation}
\label{eq:optimal-quaternion-crit}
\min_{\qvect} Q(\qvect) = \min_{\qvect} \big(  \qvect\tp \Mmat \qvect + \lambda (1- \qvect\tp\qvect) \big).
\end{equation}
From $dQ/d\qvect =0$ we obtain $\Mmat \qvect = \lambda \qvect$ and by substitution in \eqref{eq:optimal-quaternion-crit} we obtain
$Q(\qvect) = \lambda$.
Therefore, the minimization problem \eqref{eq:optimal-quaternion-crit} is equivalent to estimating the smallest eigenvalue-eigenvector pair $(\lambda^{\star}, \qvect^{\star})$ of $\Mmat$.
\section{Implementation Details}
\label{app:implementation}
Algorithm~\ref{algo:em-robfa} and  Algorithm~\ref{algo:em-student} are expectation maximization (EM) procedures and it is well known that they have good convergence properties. One should notice that all the computations inside these algorithms are in closed-form, with the notable exception of the estimation of the rotation matrix. The latter is parameterized with a unit quaternion and it is estimated via optimization of \eqref{eq:optimization-quaternion}. The unit-quaternion parameterization of rotations, i.e. Appendix~\ref{app:rotation-unitquaternion}, has several advantages: (i)~the number of parameters to be estimated is reduced from nine to four, (ii)~the number of nonlinear constraints is reduced from seven constraints (six quadratic constraints, i.e. $\Rmat\tp\Rmat=\Imat$, and one quartic constraint, i.e. $|\Rmat|=1$) to one quadratic constraint ($\qvect^{\top} \qvect=1$), (iii)~the initialization is performed with the closed-form solution of \citep{horn1987closed} that uses a unit quaternion as well.

In practice, the constrained nonlinear optimization problem \eqref{eq:optimization-quaternion} is solved using the sequential quadratic programming method \citep{bonnans2006numerical}, more precisely a sequential least squares programming (SLSQP) solver\footnote{\url{https://docs.scipy.org/doc/scipy/reference/optimize.html}}
is used in combination with a root-finding software package \citep{kraft1988software}.  The SLSQP minimizer found at the previous EM iteration is used as an initial estimate at the current EM iteration. The closed-form method of \ref{app:rotation-unitquaternion} is used to initialize the unit-quaternion, and hence the rotation matrix, at the start of the EM algorithm.


 \bibliographystyle{natbib} 

\begin{thebibliography}{10}
\providecommand{\url}[1]{#1}
\csname url@samestyle\endcsname
\providecommand{\newblock}{\relax}
\providecommand{\bibinfo}[2]{#2}
\providecommand{\BIBentrySTDinterwordspacing}{\spaceskip=0pt\relax}
\providecommand{\BIBentryALTinterwordstretchfactor}{4}
\providecommand{\BIBentryALTinterwordspacing}{\spaceskip=\fontdimen2\font plus
\BIBentryALTinterwordstretchfactor\fontdimen3\font minus
  \fontdimen4\font\relax}
\providecommand{\BIBforeignlanguage}[2]{{%
\expandafter\ifx\csname l@#1\endcsname\relax
\typeout{** WARNING: IEEEtran.bst: No hyphenation pattern has been}%
\typeout{** loaded for the language `#1'. Using the pattern for}%
\typeout{** the default language instead.}%
\else
\language=\csname l@#1\endcsname
\fi
#2}}
\providecommand{\BIBdecl}{\relax}
\BIBdecl

\bibitem{Escalera2018computational}
S.~Escalera, X.~Baro, I.~Guyon, H.~J. Escalante, G.~Tzimiropoulos, M.~Valstar,
  M.~Pantic, J.~Cohn, and T.~Kanade, ``Special issue on the computational
  face,'' \emph{IEEE Transactions on Pattern Analysis and Machine
  Intelligence}, vol.~40, no.~11, pp. 2541--2545, Nov 2018.

\bibitem{Loy2019deep}
C.~C. Loy, X.~Liu, T.-K. Kim, F.~De~la Torre, and R.~Chellappa, ``Special issue
  on deep learning for face analysis,'' \emph{International Journal of Computer
  Vision}, vol. 127, no.~6, pp. 533--536, June 2019.

\bibitem{wang2021deep}
M.~Wang and W.~Deng, ``Deep face recognition: A survey,''
  \emph{Neurocomputing}, vol. 429, pp. 215--244, 2021.

\bibitem{wu2018facial}
Y.~Wu and Q.~Ji, ``Facial landmark detection: A literature survey,''
  \emph{International Journal of Computer Vision}, vol. 127, no.~2, pp.
  115--142, 2019.

\bibitem{deng2019joint}
J.~Deng, G.~Trigeorgis, Y.~Zhou, and S.~Zafeiriou, ``Joint multi-view face
  alignment in the wild,'' \emph{IEEE Transactions on Image Processing},
  vol.~28, no.~7, pp. 3636--3648, 2019.

\bibitem{dong2018supervision}
X.~Dong, S.-I. Yu, X.~Weng, S.-E. Wei, Y.~Yang, and Y.~Sheikh,
  ``Supervision-by-registration: An unsupervised approach to improve the
  precision of facial landmark detectors,'' in \emph{Proceedings of the IEEE
  Conference on Computer Vision and Pattern Recognition}, 2018, pp. 360--368.

\bibitem{dong2020supervision}
X.~Dong, Y.~Yang, S.-E. Wei, X.~Weng, Y.~Sheikh, and S.-I. Yu, ``Supervision by
  registration and triangulation for landmark detection,'' \emph{IEEE
  Transactions on Pattern Analysis and Machine Intelligence}, vol.~43, no.~10,
  pp. 3681--3694, 2020.

\bibitem{wan2021arobust}
J.~Wan, Z.~Lai, J.~Li, J.~Zhou, and C.~Gao, ``Robust facial landmark detection
  by multiorder multiconstraint deep networks,'' \emph{IEEE Transactions on
  Neural Networks and Learning Systems}, vol.~33, no.~5, pp. 2181--2194, 2021.

\bibitem{wan2021brobust}
J.~Wan, Z.~Lai, J.~Liu, J.~Zhou, and C.~Gao, ``Robust face alignment by
  multi-order high-precision hourglass network,'' \emph{IEEE Transactions on
  Image Processing}, vol.~30, pp. 121--133, 2021.

\bibitem{wan2023precise}
J.~Wan, J.~Liu, J.~Zhou, Z.~Lai, L.~Shen, H.~Sun, P.~Xiong, and W.~Min,
  ``Precise facial landmark detection by reference heatmap transformer,''
  \emph{IEEE Transactions on Image Processing}, vol.~32, pp. 1966--1977, 2023.

\bibitem{gou2016shape}
C.~Gou, Y.~Wu, F.-Y. Wang, and Q.~Ji, ``{Shape augmented regression for 3D face
  alignment},'' in \emph{European Conference on Computer Vision}, 2016, pp.
  604--615.

\bibitem{zhu2016face}
X.~Zhu, Z.~Lei, X.~Liu, H.~Shi, and S.~Z. Li, ``{Face alignment across large
  poses: a 3D solution},'' in \emph{IEEE Conference on Computer Vision and
  Pattern Recognition}, 2016, pp. 146--155.

\bibitem{deng2019menpo}
J.~Deng, A.~Roussos, G.~Chrysos, E.~Ververas, I.~Kotsia, J.~Shen, and
  S.~Zafeiriou, ``{The Menpo benchmark for multi-pose 2D and 3D facial landmark
  localisation and tracking},'' \emph{International Journal of Computer
  Vision}, vol. 127, no. 6-7, pp. 599--624, 2019.

\bibitem{kendall2009shape}
D.~G. Kendall, D.~Barden, T.~K. Carne, and H.~Le, \emph{Shape and shape
  theory}.\hskip 1em plus 0.5em minus 0.4em\relax John Wiley \& Sons, 2009.

\bibitem{3DFA1}
A.~Bulat, ``{3D-FAN (V1.1.1)},''
  \url{https://github.com/1adrianb/face-alignment}, 2020.

\bibitem{3DFA2}
F.~Yao, ``{PRNet},'' \url{https://github.com/YadiraF/PRNet}, 2018.

\bibitem{3DFA3}
J.~Guo, ``{3DDFA},'' \url{https://github.com/cleardusk/3DDFA}, 2019.

\bibitem{3DFA4}
X.~Tu and Y.~Luo, ``{2DASL},'' \url{https://github.com/XgTu/2DASL}, 2019.

\bibitem{3DFA5}
J.~Guo, ``{3DDFA-V2},'' \url{https://github.com/cleardusk/3DDFA_V2}, 2021.

\bibitem{bulat2016two}
A.~Bulat and G.~Tzimiropoulos, ``{Two-stage convolutional part heatmap
  regression for the 1st 3D face alignment in the wild (3DFAW) challenge},'' in
  \emph{European Conference on Computer Vision Workshops}.\hskip 1em plus 0.5em
  minus 0.4em\relax Springer, 2016, pp. 616--624.

\bibitem{feng2018joint}
Y.~Feng, F.~Wu, X.~Shao, Y.~Wang, and X.~Zhou, ``{Joint 3D face reconstruction
  and dense alignment with position map regression network},'' in
  \emph{European Conference on Computer Vision}, 2018, pp. 534--551.

\bibitem{tu2dasl19}
X.~Tu, J.~Zhao, Z.~Jiang, Y.~Luo, M.~Xie, Y.~Zhao, L.~He, Z.~Ma, and J.~Feng,
  ``3{D} face reconstruction from a single image assisted by 2{D} face images
  in the wild,'' \emph{IEEE Transactions on Multimedia}, vol.~23, pp.
  1160--1172, May 2021.

\bibitem{guo2020towards}
J.~Guo, X.~Zhu, Y.~Yang, F.~Yang, Z.~Lei, and S.~Z. Li, ``Towards fast,
  accurate and stable {3D} dense face alignment,'' in \emph{European Conference
  on Computer Vision}.\hskip 1em plus 0.5em minus 0.4em\relax Springer, 2020,
  pp. 152--168.

\bibitem{UPA-3DFA}
S.~Mostafa, ``{UPA-3DFA},'' \url{https://gitlab.inria.fr/smostafa/upa3dfa},
  2023.

\bibitem{szeptycki2009coarse}
P.~Szeptycki, M.~Ardabilian, and L.~Chen, ``A coarse-to-fine curvature
  analysis-based rotation invariant {3D} face landmarking,'' in
  \emph{International Conference on Biometrics: Theory, Applications, and
  Systems}.\hskip 1em plus 0.5em minus 0.4em\relax IEEE, 2009, pp. 1--6.

\bibitem{zhu2019face}
X.~Zhu, X.~Liu, Z.~Lei, and S.~Z. Li, ``Face alignment in full pose range: A 3d
  total solution,'' \emph{IEEE Transactions on Pattern Analysis and Machine
  Intelligence}, vol.~41, no.~1, pp. 78--92, 2019.

\bibitem{ning2020real}
X.~Ning, P.~Duan, W.~Li, and S.~Zhang, ``Real-time {3D} face alignment using an
  encoder-decoder network with an efficient deconvolution layer,'' \emph{IEEE
  Signal Processing Letters}, vol.~27, pp. 1944--1948, 2020.

\bibitem{hoang20213d}
V.-T. Hoang, D.-S. Huang, and K.-H. Jo, ``{3-D} facial landmarks detection for
  intelligent video systems,'' \emph{IEEE Transactions on Industrial
  Informatics}, vol.~17, no.~1, pp. 578--586, 2021.

\bibitem{phillips2005overview}
P.~J. Phillips, P.~J. Flynn, T.~Scruggs, K.~W. Bowyer, J.~Chang, K.~Hoffman,
  J.~Marques, J.~Min, and W.~Worek, ``Overview of the face recognition grand
  challenge,'' in \emph{Conference on Computer Vision and Pattern Recognition},
  vol.~1.\hskip 1em plus 0.5em minus 0.4em\relax IEEE, 2005, pp. 947--954.

\bibitem{jeni2016first}
L.~A. Jeni, S.~Tulyakov, L.~Yin, N.~Sebe, and J.~F. Cohn, ``{The first 3D face
  alignment in the wild (3DFAW) challenge},'' in \emph{European Conference on
  Computer Vision}.\hskip 1em plus 0.5em minus 0.4em\relax Springer, 2016, pp.
  511--520.

\bibitem{sanyal2019learning}
S.~Sanyal, T.~Bolkart, H.~Feng, and M.~J. Black, ``{Learning to regress 3D face
  shape and expression from an image without 3D supervision},'' in \emph{IEEE
  Conference on Computer Vision and Pattern Recognition}, 2019, pp. 7763--7772.

\bibitem{yin126high}
L.~Yin, X.~C.~Y. Sun, T.~Worm, and M.~Reale, ``{A high-resolution 3D dynamic
  facial expression database},'' in \emph{IEEE International Conference on
  Automatic Face and Gesture Recognition}, 2008.

\bibitem{zhang2014bp4d}
X.~Zhang, L.~Yin, J.~F. Cohn, S.~Canavan, M.~Reale, A.~Horowitz, P.~Liu, and
  J.~M. Girard, ``{BP4D-spontaneous: a high-resolution spontaneous 3D dynamic
  facial expression database},'' \emph{Image and Vision Computing}, vol.~32,
  no.~10, pp. 692--706, 2014.

\bibitem{gross2010multi}
R.~Gross, I.~Matthews, J.~Cohn, T.~Kanade, and S.~Baker, ``{Multi-PIE},''
  \emph{Image and Vision Computing}, vol.~28, no.~5, pp. 807--813, 2010.

\bibitem{jeni2017dense}
L.~A. Jeni, J.~F. Cohn, and T.~Kanade, ``{Dense 3D face alignment from 2D video
  for real-time use},'' \emph{Image and Vision Computing}, vol.~58, pp. 13--24,
  2017.

\bibitem{bagdanov2011florence}
A.~D. Bagdanov, A.~Del~Bimbo, and I.~Masi, ``{The Florence 2D/3D hybrid face
  dataset},'' in \emph{Joint ACM Workshop on Human Gesture and Behavior
  Understanding}, 2011, pp. 79--80.

\bibitem{McLachlanPeel2000b}
G.~McLachlan and D.~Peel, ``Robust mixture modelling using the t
  distribution,'' \emph{Statistics and Computing}, vol.~10, no.~4, pp.
  339--348, 2000.

\bibitem{sun2010robust}
J.~Sun, A.~Kab{\'a}n, and J.~M. Garibaldi, ``Robust mixture clustering using
  {P}earson type {VII} distribution,'' \emph{Pattern Recognition Letters},
  vol.~31, no.~16, pp. 2447--2454, 2010.

\bibitem{forbes2014new}
F.~Forbes and D.~Wraith, ``A new family of multivariate heavy-tailed
  distributions with variable marginal amounts of tailweight: application to
  robust clustering,'' \emph{Statistics and computing}, vol.~24, no.~6, pp.
  971--984, 2014.

\bibitem{chamroukhi2017skew}
F.~Chamroukhi, ``Skew t mixture of experts,'' \emph{Neurocomputing}, vol. 266,
  pp. 390--408, 2017.

\bibitem{banfield1993model}
J.~D. Banfield and A.~E. Raftery, ``Model-based gaussian and non-gaussian
  clustering,'' \emph{Biometrics}, pp. 803--821, 1993.

\bibitem{zaharescu2009robust}
A.~Zaharescu and R.~Horaud, ``{Robust factorization methods using a
  Gaussian/uniform mixture model},'' \emph{International Journal of Computer
  Vision}, vol.~81, no.~3, pp. 240--258, 2009.

\bibitem{myronenko2010point}
A.~Myronenko and X.~Song, ``Point set registration: Coherent point drift,''
  \emph{IEEE Transactions on Pattern Analysis and Machine Intelligence},
  vol.~32, no.~12, pp. 2262--2275, 2010.

\bibitem{lathuiliere2018deepgum}
S.~Lathuili{\`e}re, P.~Mesejo, X.~Alameda-Pineda, and R.~Horaud, ``{DeepGUM}:
  Learning deep robust regression with a {Gaussian}-uniform mixture model,'' in
  \emph{European Conference on Computer Vision}, 2018, pp. 202--217.

\bibitem{horn1987closed}
B.~K. Horn, ``Closed-form solution of absolute orientation using unit
  quaternions,'' \emph{Journal of the Optical Society of America A}, vol.~4,
  no.~4, pp. 629--642, 1987.

\bibitem{Umeyama91}
S.~Umeyama, ``Least-squares estimation of transformation parameters between two
  point patterns,'' \emph{IEEE Transactions on Pattern Analysis and Machine
  Intelligence}, vol.~13, no.~4, pp. 376--380, April 1991.

\bibitem{viola2001rapid}
P.~Viola and M.~Jones, ``Rapid object detection using a boosted cascade of
  simple features,'' in \emph{Computer Vision and Pattern Recognition},
  vol.~1.\hskip 1em plus 0.5em minus 0.4em\relax Ieee, 2001.

\bibitem{drouard2017robust}
V.~Drouard, R.~Horaud, A.~Deleforge, S.~Ba, and G.~Evangelidis, ``Robust
  head-pose estimation based on partially-latent mixture of linear
  regressions,'' \emph{IEEE Transactions on Image Processing}, vol.~26, no.~3,
  pp. 1428--1440, 2017.

\bibitem{abtahi2014yawdd}
S.~Abtahi, M.~Omidyeganeh, S.~Shirmohammadi, and B.~Hariri, ``{YawDD: a yawning
  detection dataset},'' in \emph{ACM Multimedia Systems Conference}, 2014, pp.
  24--28.

\bibitem{savage1972foundations}
L.~J. Savage, \emph{The Foundations of Statistics}.\hskip 1em plus 0.5em minus
  0.4em\relax Dover, 1972.

\bibitem{huber2018logical}
F.~Huber, \emph{A Logical Introduction to Probability and Induction}.\hskip 1em
  plus 0.5em minus 0.4em\relax Oxford University Press, 2018.

\bibitem{bulat2017far}
A.~Bulat and G.~Tzimiropoulos, ``{How far are we from solving the 2D \& 3D face
  alignment problem? (and a dataset of 230,000 3D facial landmarks)},'' in
  \emph{IEEE International Conference on Computer Vision}, 2017, pp.
  1021--1030.

\bibitem{zhang2020deep}
J.~Zhang, K.~Gao, K.~Fu, and P.~Cheng, ``Deep {3D} facial landmark localization
  on position maps,'' \emph{Neurocomputing}, vol. 406, pp. 89--98, 2020.

\bibitem{wang2022learning}
Y.~Wang, M.~Cao, Z.~Fan, and S.~Peng, ``Learning to detect {3D} facial
  landmarks via heatmap regression with graph convolutional network,'' in
  \emph{The 36th AAAI Conference on Artificial Intelligence}, 2022.

\bibitem{akhondi2014logarithmic}
A.~Akhondi-Asl, L.~Hoyte, M.~E. Lockhart, and S.~K. Warfield, ``A logarithmic
  opinion pool based {STAPLE} algorithm for the fusion of segmentations with
  associated reliability weights,'' \emph{IEEE Transactions on Medical
  Imaging}, vol.~33, no.~10, pp. 1997--2009, 2014.

\bibitem{bonnans2006numerical}
J.-F. Bonnans, J.~C. Gilbert, C.~Lemar{\'e}chal, and C.~A. Sagastiz{\'a}bal,
  \emph{Numerical optimization: theoretical and practical aspects}.\hskip 1em
  plus 0.5em minus 0.4em\relax Springer Science \& Business Media, 2006.

\bibitem{kraft1988software}
D.~Kraft, ``A software package for sequential quadratic programming,'' DLR
  German Aerospace Center -- Institute for Flight Mechanics, Koln, Germany,
  Tech. Rep. DFVLR-FB 88-28, 1988.

\end{thebibliography}

\end{document}